%% file: main.tex
\RequirePackage[hyphens]{url}
\documentclass{article}
\usepackage{authblk}
\usepackage[margin=1.4in]{geometry}

\input macros.tex
\title{DeepSpeed-MoE: Advancing Mixture-of-Experts Inference and Training to Power Next-Generation AI Scale}
\author[]{Samyam Rajbhandari}
\author[]{Conglong Li}
\author[]{Zhewei Yao}
\author[]{Minjia Zhang}
\author[]{Reza Yazdani Aminabadi}
\author[]{Ammar Ahmad Awan}
\author[]{Jeff Rasley}
\author[]{Yuxiong He}
\affil[]{Microsoft}
\date{}
\usepackage{xspace}
\usepackage{graphicx}
\usepackage{multirow}
\usepackage{hhline}
\usepackage{comment}
\usepackage{adjustbox}
\usepackage{xcolor}
\usepackage{pifont}
\usepackage{subcaption}

\usepackage{tabulary}
\usepackage{array}
\usepackage{colortbl}

\usepackage{color,soul}
\usepackage[bookmarks=true,breaklinks=true,colorlinks,linkcolor=black,citecolor=blue,urlcolor=black]{hyperref}

\renewcommand{\baselinestretch}{1.0}
\usepackage[switch,columnwise]{lineno} 

\usepackage{amsmath}
\usepackage{amsfonts}       

\newcommand{\pyramidmoe}{Pyramid-MoE\xspace}
\newcommand{\residualmoe}{Residual-MoE\xspace}
\newcommand{\prmoe}{PR-MoE\xspace}
\newcommand{\standardmoe}{Standard-MoE\xspace}

\newcommand\extrafootertext[1]{%
    \bgroup
    \renewcommand\thefootnote{\fnsymbol{footnote}}%
    \renewcommand\thempfootnote{\fnsymbol{mpfootnote}}%
    \footnotetext[0]{#1}%
    \egroup
}

\begin{document}

\maketitle
\input{_s0_abstract}

\extrafootertext{This paper is published at ICML 2022~\cite{pmlr-v162-rajbhandari22a}: \url{https://proceedings.mlr.press/v162/rajbhandari22a}. Author contributions listed at \hyperref[sec:contribution]{the end of paper.}}

\input{_s1_introduction}

\input{_s2_related_work}

\input{_s3_standardmoe}

\input{_s4_1_prmoe}

\input{_s5_1_inference-system}

\input{_s6_conclusion}

\input{_s7_ack}

\bibliographystyle{unsrt}
\bibliography{references}

\newpage
\appendix

\end{document}

%% file: macros.tex
\newcommand\fref{Figure~\ref}
\newcommand\tref{Table~\ref}
\newcommand\sref{Section~\ref}

%% file: _s0_abstract.tex
\begin{abstract}
    As the training of giant dense models hits the boundary on the availability and capability of the hardware resources today, Mixture-of-Experts (MoE) models become one of the most promising model architectures due to their significant training cost reduction compared to a quality-equivalent dense model. Its training cost saving is demonstrated from encoder-decoder models (prior works) to a 5x saving for auto-aggressive language models (this work along with parallel explorations). However, due to the much larger model size and unique architecture, how to provide fast MoE model inference remains challenging and unsolved, limiting its practical usage. To tackle this, we present DeepSpeed-MoE, an end-to-end MoE training and inference solution as part of the DeepSpeed library, including novel MoE architecture designs and model compression techniques that reduce MoE model size by up to 3.7x, and a highly optimized inference system that provides 7.3x better latency and cost compared to existing MoE inference solutions. DeepSpeed-MoE offers an unprecedented scale and efficiency to serve massive MoE models with up to 4.5x faster and 9x cheaper inference compared to quality-equivalent dense models. We hope our innovations and systems help open a promising path to new directions in the large model landscape, a shift from dense to sparse MoE models, where training and deploying higher-quality models with fewer resources becomes more widely possible.
\end{abstract}

%% file: _s1_introduction.tex
\section{Introduction}

In the last three years, the largest trained model has increased in size by over 1000x, from a few hundred million parameters to half a trillion parameters (Megatron-Turing NLG 530B). 
Improvements in model quality with size suggest that this trend will continue, with larger model sizes bringing better model quality. 

However, sustaining the growth in model size is getting more and more difficult due to the increasing compute requirements.
For example, the largest single dense model in existence as of Dec 2021, the Megatron-Turing NLG 530B model, took around 3 months to train on over 2000 A100 GPUs on the NVIDIA Selene Supercomputer, consuming over 3 million GPU hours~\cite{mt-nlg}.
Another 3 to 5 times of increase in dense model size would be infeasible within a reasonable timeframe.  

Given the exorbitant compute resources required to train the state-of-art models, a natural question to ask is: ``Is it possible to make non-trivial improvement to model quality without increasing the compute cost?''  
Or equivalently, ``Is it possible to produce model with similar quality using 3 to 5 times less resources?''

There have been numerous efforts to reduce the compute requirements to train large models without sacrificing model quality. 
To this end, architectures based on Mixture-of-Experts (MoE)~\cite{shazeer2017outrageously, lepikhin2020gshard, fedus2021switch} have paved a promising path, enabling sub-linear compute requirements with respect to the model parameters and allowing for improved model quality without increasing training cost. 
However, MoE based models have their own set of challenges that limit their use in a wide range of real world scenarios:
\begin{itemize}
    \item \textbf{Limited Scope} The scope of MoE based models in the NLP area is primarily limited to encoder-decoder models and sequence-to-sequence tasks, with limited work done in exploring its application in other domains. 
    Application of MoE to auto-regressive natural language generation (NLG) like GPT-3 and MT-NLG 530B, where the compute cost of training state-of-art language models can be orders of magnitude higher than for encoder-decoder models, is less explored.
    
    \item \textbf{Massive Memory Requirements} While MoE models require less compute to achieve the same model quality as their dense counterparts, they need significantly more number of parameters.
    For example, the MoE based Switch-Base model has 10x more parameters than T5-large (7.4B vs 0.74B) and still it does not have the same model quality when compared across a wide range of downstream tasks~\cite{fedus2021switch}. 
    In other words, MoE based models have a much lower ``parameter efficiency'' compared to quality-equivalent dense models. Larger model size and lower parameter efficiency bring challenges in both training and inference.
    
 For training, this massive increase in model size requires a proportional increase in device memory. 
    Note that the above mentioned T5-large (0.74B) can fit in the memory of a single 32GB V100 GPU, while training Switch-Base (7.4B) requires at least 8-10 such GPUs to even just fit the model in device memory for training. 
    If we scale the dense model size to MT-NLG equivalent with 500B parameters, achieving similar quality with MoE based model might need a model with over 5 trillion parameters (assuming the 10x scaling still holds), which would require over 5K GPUs to just fit the model states for training. 
    This massive model size makes MoE based model training at scale challenging due to both the device memory required as well as the system support required to effectively utilize the device memory across thousands of GPUs.

    \item \textbf{Limited Inference Performance} Due to the large model size and poor parameter efficiency mentioned above, fast inference of MoE based models is even more challenging.
    On one hand, the larger parameter size requires more GPUs to fit, and multi-gpu inference technology is not designed to work with MoE based models. 
    On the other hand, as inference is often memory bandwidth bound, MoE based models, which can be 10x larger than their dense equivalent, could require 10x higher achievable memory bandwidth to achieve similar inference latency as the dense models. 
\end{itemize}

Despite the promising and non-trivial reduction in training cost, these above mentioned challenges severely limits the practical applicability of MoE. 
In an effort to make MoE practical, accessible and applicable, in this paper, we address these challenges by offering three corresponding solutions:

\begin{itemize}
\item We expand the scope of MoE based models to auto-regressive NLG tasks, demonstrating training cost reduction of 5x to achieve same model quality for models like GPT-3 and MT-NLG. 
These results not only demonstrate clear opportunities to reduce the cost of training massive NLG models, but also opens up the possibilities to reach much higher next-generation model quality under the limitation of current generation hardware resource.

\item We improve parameter efficiency of MoE based models by developing a novel MoE architecture that we call Pyramid-Residual MoE (PR-MoE). 
PR-MoE is a hybrid dense and MoE model created using residual connections, while applying experts only where they are most effective. 
PR-MoE can reduce MoE model parameter size by up to 3x with no change to model quality and minimal change to the compute requirements. 
In addition, we create a distilled version of PR-MoE, which we call Mixture-of-Students (MoS), via staged knowledge distillation. MoS reduces the MoE model size by up to 3.7x while retaining comparable model quality.

\item  We develop DeepSpeed-MoE inference system, a highly optimized MoE inference system which enables efficient scaling of inference workloads on hundreds of GPUs, providing up to 7.3x reduction in inference latency and cost when compared with existing MoE inference solutions. It offers ultra-fast inference latencies (under 25 ms) for trillion-parameter MoE models. DeepSpeed-MoE also offers up to 4.5x faster and 9x cheaper inference for MoE models compared to quality-equivalent dense models by combining both system and model optimizations. 
\end{itemize}

Together, our innovations and systems enable MoE to be a more effective and economic alternative comparing to dense models, achieving significantly lower training and inference cost while obtaining the same model quality.  Or equivalently, they can be used to empower the models of the next generation AI scale without requiring an increase in compute resources.  
We hope DeepSpeed-MoE helps open a promising path to new directions in the large model landscape, a shift from dense to sparse MoE models, where training and deploying higher-quality models with fewer resources becomes more widely possible.

\paragraph{Paper outline} We begin the remainder of this paper with necessary background and related work followed by the three solutions above presented as self-contained sections. 

\paragraph{Software} We are in the process of open sourcing DeepSpeed-MoE as part of the DeepSpeed library over multiple stages.
Please find the code, tutorials, and documents at DeepSpeed GitHub (\url{https://github.com/microsoft/DeepSpeed}) and website (\url{https://www.deepspeed.ai/}). Experiments in this paper were conducted on the Microsoft Azure AI platform, the best place to train and serve models using DeepSpeed. Our tutorials include how to get started with DeepSpeed on Azure and experiment with different models using Azure ML.

%% file: _s2_related_work.tex
\section{Related Work}

\subsection{Large Scale Dense NLP Models} To test and verify the upper bound of scaling law~\cite{kaplan2020scaling} for model capacity with respect to number of parameters, the pretrained natural language processing model size has been increasing 10x per year for the last several years. 
Earlier works are generally in the scale of hundreds of millions of parameters, including BERT~\cite{devlin2019bert}, XLNet~\cite{yang2019xlnet}, RoBERTa~\cite{liu2019roberta}, ALBERT~\cite{lan2019albert}, and GPT~\cite{radford2018improving}, etc. 
Later on, billions to dozens of billions models, like GPT-2~\cite{radford2019language}, TuringNLG~\cite{rosset2020turing}, Megatron-LM~\cite{shoeybi2019megatron}, T5~\cite{raffel2019exploring} etc, are introduced and they are shown to have better generalization performance on various of natural language understanding and generation tasks~\cite{paperno2016lambada,wang2018glue,wang2019superglue,mostafazadeh2016corpus,berant2013semantic,joshi2017triviaqa}. 
The GPT-3~\cite{brown2020language} further pushes the upper limit to 175 billions parameters, and shows that with zero/few-shot learning, it can achieve comparable or even better performance than previous small scale models with finetuning.
More recently, the extra-large Megatron-Turing NLG 530B model~\cite{mt-nlg} achieves 530 billions parameters by software support from DeepSpeed~\cite{rajbhandari2020zero} and Megatron-LM~\cite{shoeybi2019megatron}, and it achieves new state-of-the-art results of zero/few-shot learning within one dense model. 
However, as the training takes 3 months on over 2000 A100 GPUs, it is no longer feasible to achieve better model quality by simply increasing the model size due to unsurmountable compute requirements. 

\subsection{Reducing Training Cost by MoE Architecture} One promising way to reduce the training cost is using Mixture of Expert (MoE)~\cite{masoudnia2014mixture}.
In \cite{shazeer2017outrageously} authors scale LSTM based model to 127B by applying convolutional MoE layers between stacked LSTM layers~\cite{hochreiter1997long} for language modelings and machine translation.
The sparsity nature of MoE significantly improves the scaling of model size without increasing the computational cost. 
GShard~\cite{lepikhin2020gshard} utilizes MoE to train a transformer-based model~\cite{vaswani2017attention} to 600B parameters for multi-language translation, and it shows that the training cost of this 600B MoE model is even cheaper than that of a 100B dense model. 
Switch Transformer~\cite{fedus2021switch} continues this based on the T5 model and scales the model to 1.6 trillion. 
To achieve same accuracy performance, \cite{fedus2021switch} shows a 2.5x faster training speed of MoE models as compared to large dense models.
Following by this, more recent work~\cite{lin2021m6,kim2021scalable,zuo2021taming} exhibit the sparsity advantage of MoE.

A few recent and parallel works~\cite{google_glam,artetxe2021efficient} show that MoE model can also be applied to auto-regressive natural language generation tasks, including preliminary results on multiple downstream evaluation tasks. 
However, our work has major differences compared to these parallel explorations:
(1) our work investigates training, model design, and inference opportunities of MoE models while~\cite{google_glam,artetxe2021efficient} primarily focuses on MoE training;
(2) we propose PR-MoE architecture and MoS knowledge distillation to achieve better MoE parameter efficiency and on-par/better zero-shot eval quality as described in~\sref{section_standard_moe_quality};
(3) we develop DeepSpeed-MoE inference system to efficiently serve large scale MoE models with high throughput and low latency. While recent studies like ~\cite{google_glam} and~\cite{artetxe2021efficient} discuss reduction of FLOPs, it is pertinent to mention that unlike training, inference latency and cost do not depend on computation alone. Efficient inference depends on model size, memory bandwidth, and the capability of a system to read data from memory efficiently. DeepSpeed-MoE inference system is an end-to-end system that combines model architecture innovations (Section~\ref{sec:moe_parameter_efficiency}) and myriad of optimizations and techniques (Section~\ref{sec:optimizing_moe_inference_latency}) to deliver ultra low latency and super-linear speedups for throughput.

\subsection{MoE Training and Inference Systems} Given the recent uptake of MoE models by many researchers around the world, open source systems and frameworks are being developed or extended to support MoE model training. To the best of our knowledge, there are no MoE systems specifically designed and optimized for inference yet. Several MoE training systems have been presented recently. Below we discuss a few of them. 

DeepSpeed MoE training system ~\cite{deepspeed-moe-paper} was primarily targeted for optimized training of MoE models at scale. The main goal of this work was to establish a flexible user-facing API for extending existing training codes for efficient and scalable MoE model training. It supports up to 8x bigger model sizes by leveraging flexible combinations of different types of parallelism including tensor-slicing, data parallelism, ZeRO~\cite{rajbhandari2020zero}-powered data parallelism, and expert parallelism. It was used to train Z-code MoE, a 10 billion-parameter multitask multilingual MoE model (transformer encoder-decoder architecture) on 50 languages with 5x less training time compared to a dense model of equal quality.

FastMoE~\cite{fast-moe} is a research software developed to show how MoE models can be trained under data and expert (model) parallelism. The combination of various parallelism dimensions is not fully supported. Fast-MoE examples include Transformer-XL and Megatron-LM but results about large-scale end-to-end training are not available yet. Fairseq-MoE~\cite{artetxe2021efficient} offers an MoE API as well as a training pipeline for generic language models. The Fairseq system has been further optimized by Tutel~\cite{tutel}, which offers up to 40 percent improvement over Fairseq.

%% file: _s3_standardmoe.tex
\section{DeepSpeed-MoE for NLG: Reducing the Training Cost of Language Models by 5 Times}
\label{sec:standard_moe}

Autoregressive transformer-based natural language generation (NLG) models offer convincing solutions to a broad range of language tasks from document summarization, headline generation, question and answering to even generating code in a wide variety of programming languages. 
Due to the general applicability of these models, improving their quality has been of great interest for both academia and industry alike. 
Given the tremendous compute and energy requirements for training NLG family of models, we explore the opportunities that MoE presents to reduce their training cost. 
We show that MoE can be applied to NLG family of models to significantly improve their model quality with the same training cost. 
Alternatively, it can achieve 5x reduction in training cost to achieve the same model quality of a dense NLG model. 

\subsection{MoE based NLG Model Architecture}
To create an MoE based NLG model, we studied the GPT like transformer-based NLG model. 
To complete training in a reasonable timeframe, the following models are selected: 350M (24 layers, 1024 hidden size, 16 attention heads), 1.3B (24 layers, 2048 hidden size, 16 attention heads), and 6.7B (32 layers, 4096 hidden size, 32 attention heads). 
We use “350M+MoE-128” to denote a MoE model that uses 350M dense model as the base model and adds 128 experts on every other feedforward layer. 
That is to say, there are in total 12 MoE layers for both 350M+MoE-128 and 1.3B+MoE-128.

We use a gating function to activate a subset of experts in the MoE layer for each token. 
Specifically, in our experiments, only the top-1 expert is selected. Therefore, during both training and inference, our MoE model will have the same number of parameters to be activated for each token as their dense part (illustrated in~\fref{fig:standard_pr_moe} (left)). 
For example, 1.3B+MoE-128 will only activate 1.3B parameter per token, and the amount of training computation per token will be similar to a 1.3B dense model. We also tested top-2 gating function and found it provides some convergence improvement, but it is a diminishing return and it comes with a large training computation/communication overhead compared to top-1 gating.

\subsection{Training and Evaluation Settings}
We pre-trained both the dense and MoE version of the above models using DeepSpeed on 128 Ampere A100 GPUs (Azure ND A100 instances). 
These Azure instances are powered by the latest Azure HPC docker images that provide a fully optimized environment and best performing library versions of NCCL, Mellanox OFED, Sharp, and CUDA.
DeepSpeed uses a combination of data parallel and expert parallel training to effectively scale the MoE model training. 
We used the same training data for the MT-NLG model~\cite{mt-nlg}. 
For a fair comparison, we use 300B tokens to train both the dense model and the MoE models.

\tref{table_hparams_dense_standardmoe} summarizes the hyperparameters for training the dense and MoE models. 
For dense models we followed the hyperparameters from the GPT-3 work~\cite{brown2020language}. 
For MoE models, we find that using a lower learning rate and longer learning rate decay duration compared to the dense counter parts (e.g., dense 1.3B versus 1.3B+MoE-128) could provide better convergence. 
We believe that this is because MoE models have much larger number of parameters. 
MoE models have two additional hyperparameters: the number of experts per MoE layer, and a coefficient when adding the MoE layer losses to the total training loss.

In addition to comparing the validation loss during pre-training, we employ 6 zero-shot evaluation tasks to compare the final model quality: 1 completion prediction task (LAMBADA~\cite{paperno2016lambada}), 1 common sense reasoning task (PIQA~\cite{bisk2020piqa}), 2 reading comprehension tasks (BoolQ~\cite{wang2019superglue}, RACE-h~\cite{lai2017race}), and 2 question answering tasks (TriviaQA~\cite{joshi2017triviaqa}, WebQs~\cite{berant2013semantic}).

\begin{table*}[t]
\centering
  \footnotesize
  \caption{Hyperparameters for different dense and MoE NLG models.}\label{table_hparams_dense_standardmoe}
  \begin{tabular}{rrrrrrrrrrrrrrrr}
  \hline
  & Dense & Dense & Dense & 350M+ & 1.3B+   &350M+PR-   &1.3B+PR-\\
  & 350M & 1.3B & 6.7B & MoE-128 & MoE-128  &MoE-32/64 &MoE-64/128  \\
  \hline
  Num. layers & 24 & 24 & 32 & 24 & 24 & 24 & 24\\
  Hidden size & 1024 & 2048 & 4096 & 1024 & 2048 & 1024 & 2048\\
  Num. attention heads & 16 & 16 & 32 & 16 & 16 & 16 & 16\\
  Num. experts per layer & N/A & N/A & N/A & 128 & 128 & 32/64 & 64/128\\
  Num. parameters & 350M & 1.3B & 6.7B & 13B & 52B & 4B & 31B\\
  Context/sequence length & 2K & 2K & 2K & 2K & 2K & 2K & 2K\\
  Training tokens & 300B & 300B & 300B & 300B & 300B & 300B & 300B\\
  Batch size & 256 & 512 & 1024 & 256 & 512 & 256 & 512\\
  Batch size rampup tokens & 0B & 4B & 10B & 0B & 0B & 0B & 0B \\
  Learning rate & 3.0e-4 & 2.0e-4 & 1.2e-4 & 2.0e-4 & 1.2e-4 & 3.0e-4 & 1.2e-4\\
  Min. learning rate & 3.0e-5 & 2.0e-5 & 1.2e-5 & 2.0e-6  & 1.0e-6 & 1.0e-6 & 1.0e-6\\
  LR linear warmup tokens & 375M & 375M & 375M & 375M & 375M & 375M & 375M\\
  LR cosine decay tokens & 260B & 260B & 260B & 300B & 300B & 300B & 300B\\
  Model parallel degree & 1 & 1 & 8 & 1 & 1 & 1 & 1\\
  MoE loss coefficient & N/A & N/A & N/A & 0.01  & 0.01 & 0.01 & 0.01\\
  \hline
  \end{tabular}\vspace{-0.2cm}
\end{table*}

\subsection{MoE Leads to Better Quality for NLG Models}
\label{section_standard_moe_quality}
\begin{figure}[t]
\centering
\includegraphics[width=0.5\textwidth]{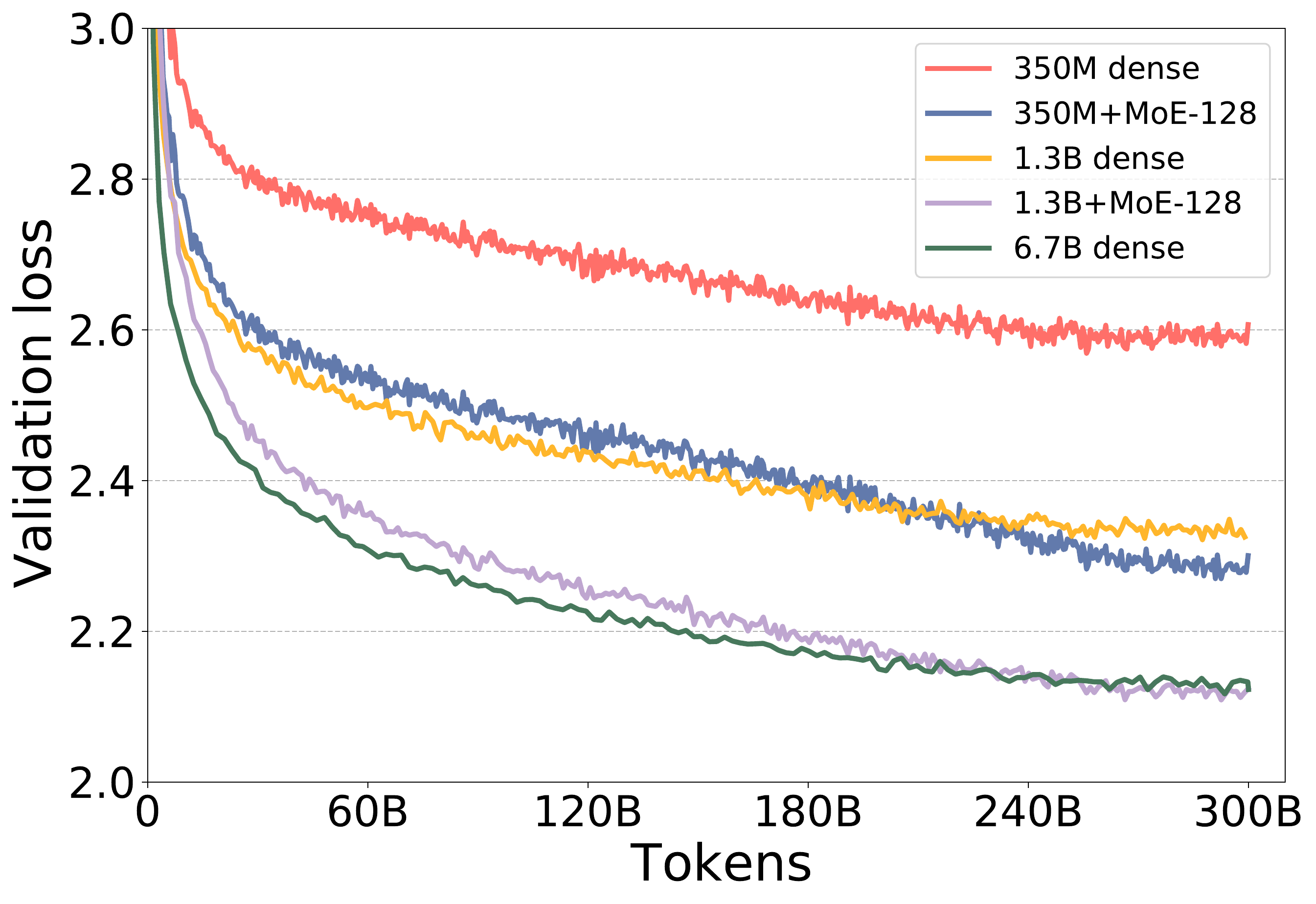}
\caption{Token-wise validation loss curves for dense and MoE NLG models with different model sizes.}
\label{figure-valid-loss}\vspace{-0.2cm}
\end{figure}

\fref{figure-valid-loss} shows that the validation loss for the MoE versions of the model is significantly better than their dense counter parts. 
Furthermore, notice that the validation loss of the MoE model, 350M+MoE-128, is on par with the validation loss of the 1.3B dense model with 4x larger base. 
This is also true for 1.3B+MoE-128 in comparison with 6.7B dense model with 5x larger base. 
Furthermore, the model quality is on par not only for the validation loss but also for the zero-shot evaluation on the 6 downstream tasks as shown in~\tref{table_zero_shot_eval}, demonstrating that MoE models and their dense counter part with 4-5x larger base have very similar model quality. 

In addition, in~\tref{table_zero_shot_eval} we also compare our zero-shot results with related works that explored MoE for NLG models in parallel with ours~\cite{google_glam, artetxe2021efficient}. 
Although with the caveats that the training data and hyperparamters are not the same, the comparisons demonstrate that on certain tasks our MoE models are able to achieve on-par or better quality with less number of parameters: Compared to 8B+MoE-64 (143B) in~\cite{google_glam}, our MoE models (1.3B+MoE-128 (52B), 1.3B+PR-MoE-64/128 (31B), 1.3B+PR-MoE+L21+MoS (27B)) are able to achieve better LAMBADA accuracy with up to 5.3x less number of parameters. Compared to the 355M+MoE-512 (52B) in~\cite{artetxe2021efficient}, our MoE models (1.3B+PR-MoE-64/128 (31B) and 1.3B+PR-MoE+L21+MoS (27B)) are able to achieve better PIQA/BoolQ accuracy with up to 1.9x less number of parameters.

\begin{table*}[t]
\centering
  \footnotesize
  \caption{Zero-shot evaluation results (last six columns) for different dense and MoE NLG models. 
  All zero-shot evaluation results use the accuracy metric. Details about results of PR-MoE NLG and Mixture-of-Students NLG are described in Section~\ref{sec:moe_parameter_efficiency}.
  Last 6 rows are related works from Google~\cite{google_glam} and Meta (Facebook)~\cite{artetxe2021efficient}.}
  \label{table_zero_shot_eval}
  \begin{tabular}{rcccccccccccccc}
  \hline
  Model (num. params) & LAMBADA & PIQA & BoolQ & RACE-h & TriviaQA & WebQs\\
  \hline
  \textbf{Dense NLG:} &  &  &  &  & & \\
  350M (350M) & 52.03 & 69.31 & 53.64 & 31.77 & 3.21 & 1.57 \\
  1.3B (1.3B) & 63.65 & 73.39 & 63.39 & 35.60 & 10.05 & 3.25 \\
  6.7B (6.7B) & 71.94 & 76.71 & 67.03 & 37.42 & 23.47 & 5.12 \\
  \hline
  \textbf{Standard MoE NLG:} &  &  &  &  & & \\
  350M+MoE-128 (13B) & 62.70 & 74.59 & 60.46 & 35.60 & 16.58 & 5.17 \\
  1.3B+MoE-128 (52B) & 69.84 & 76.71 & 64.92 & 38.09 & 31.29 & 7.19 \\
  \hline
  \textbf{PR-MoE NLG:} &  &  &  &  & & \\
  350M+PR-MoE-32/64 (4B) & 63.65 & 73.99 & 59.88 & 35.69 & 16.30 & 4.73 \\
  1.3B+PR-MoE-64/128 (31B) & 70.60	& 77.75 & 67.16 & 38.09 & 28.86 & 7.73 \\
  \hline
  \textbf{Mixture-of-Students NLG:}  &  &  &  &  & & \\
  350M+PR-MoE+L21+MoS ({3.5B}) & 63.46 & 73.34 & 58.07 & 34.83 & {13.69} & 5.22 \\
  1.3B+PR-MoE+L21+MoS ({27B}) & 70.17 & 77.69 & 65.66 & 36.94 & 29.05 & 8.22 \\
  \hline
  \textbf{Related MoE works:} &  &  &  &  & & \\
  0.1B+MoE-64 (1.9B)~\cite{google_glam} & 36.9 & 69.0 & 53.6 & 29.1 & 15.2 & 5.9 \\
  1.7B+MoE-64 (27B)~\cite{google_glam} & 63.7 & 76.6 & 64.4 & 40.7 & 42.0 & 8.5 \\
  8B+MoE-64 (143B)~\cite{google_glam} & 67.3 & 78.6 & 72.2 & 43.4 & 55.1 & 10.7 \\
  125M+MoE-512 (15B)~\cite{artetxe2021efficient} & N/A & 74.3 & 60.9 & N/A & N/A & N/A \\
  355M+MoE-512 (52B)~\cite{artetxe2021efficient} & N/A & 76.8 & 56.0 & N/A & N/A & N/A \\
  1.3B+MoE-512 (207B)~\cite{artetxe2021efficient} & N/A & 78.2 & 54.2 & N/A & N/A & N/A \\
  \hline
  \end{tabular}\vspace{-0.2cm}
\end{table*}

\subsection{Same Quality with 5x Less Training Cost}

As we saw from the results above, adding MoE with 128 experts to the NLG model significantly improves the quality of the NLG model. 
However, these experts do not change the compute requirements of the model as each token is only processed by a single expert. 
Therefore, the compute requirements for dense model and its corresponding MoE models with the same base are similar. 
More concretely, a 1.3B+MoE-128 model training requires roughly the same amount of compute operations as 1.3B dense model, while offering much better model quality. 

Furthermore, our results show that by applying MoE we can achieve the model quality of a 6.7B parameter dense model at the training cost of 1.3B parameter dense model, resulting in an effective training compute reduction of 5x. 
This compute cost reduction can directly be translated into throughput gain, training time and training cost reduction by leveraging the efficient DeepSpeed MoE training system. 
Table~\ref{table_training_throughput} shows the training throughput of the 1.3B+MoE-128 model in comparison to the 6.7B dense model on 128 NVIDIA A100 GPUs.

\begin{table*}[ht]
\centering
  \footnotesize
  \caption{Training throughput (on 128 A100 GPUs) comparing MoE based model vs dense model that can both achieve the same model quality.}\label{table_training_throughput}
  \begin{tabular}{rrr}
  \hline
  & Training samples per sec &	Throughput gain / Cost Reduction \\
  \hline
  6.7B dense & 70 & 1x \\
  1.3B+MoE-128 & 372 & 5x \\
  \hline
  \end{tabular}\vspace{-0.2cm}
\end{table*}

To conclude, this section shows significant training cost saving of using MoE on NLG models: by applying MoE we achieved the model quality of a 6.7B parameter dense NLG model at the cost of training a 1.3B base model, thanks to the sparse structure of MoE. 
Assuming the scaling holds, the results have the potential to transform the large model training landscape and power much bigger model scale under more affordable time and cost using the hard resources available today. 
For example, a model with comparable accuracy as trillion-parameter dense model can be potentially trained at the cost of a 200B parameter (like GPT-3) sized dense model, translating to millions of dollars in training cost reduction and energy savings~\cite{brown2020language}. 

%% file: _s4_1_prmoe.tex
\section{\prmoe and MoS: Reducing the Model Size and Improving Parameter Efficiency}
\label{sec:moe_parameter_efficiency}
While MoE based models achieve the same quality with 5x training cost reduction in the NLG example, the resulting model has roughly 8x the parameters of the corresponding dense model (e.g., 6.7B dense model has 6.7 billion parameters and 1.3B+MoE-128 has 52 billion parameters). 
Such a massive MoE model requires significantly more memory during training, and it is challenging to meet latency requirements for such models during inference as memory bandwidth consumed to read the model weights is the primary performance bottleneck in inference. 
To reduce the number of parameters and improve the parameter efficiency of MoE based models, we present innovations in the MoE model architecture (called \prmoe) that reduce the overall model size by up to 3 times without affecting model quality. 
In addition, we design a novel MoE-to-MoE knowledge distillation technique to create a distilled version of PR-MoE, which we call Mixture-of-Students (MoS), that further reduces the MoE model size, optimizing inference time and cost.  
Below we start with our new \prmoe architecture and then discuss MoS.

\begin{figure}[t]
\centering
\includegraphics[width=0.49\textwidth]{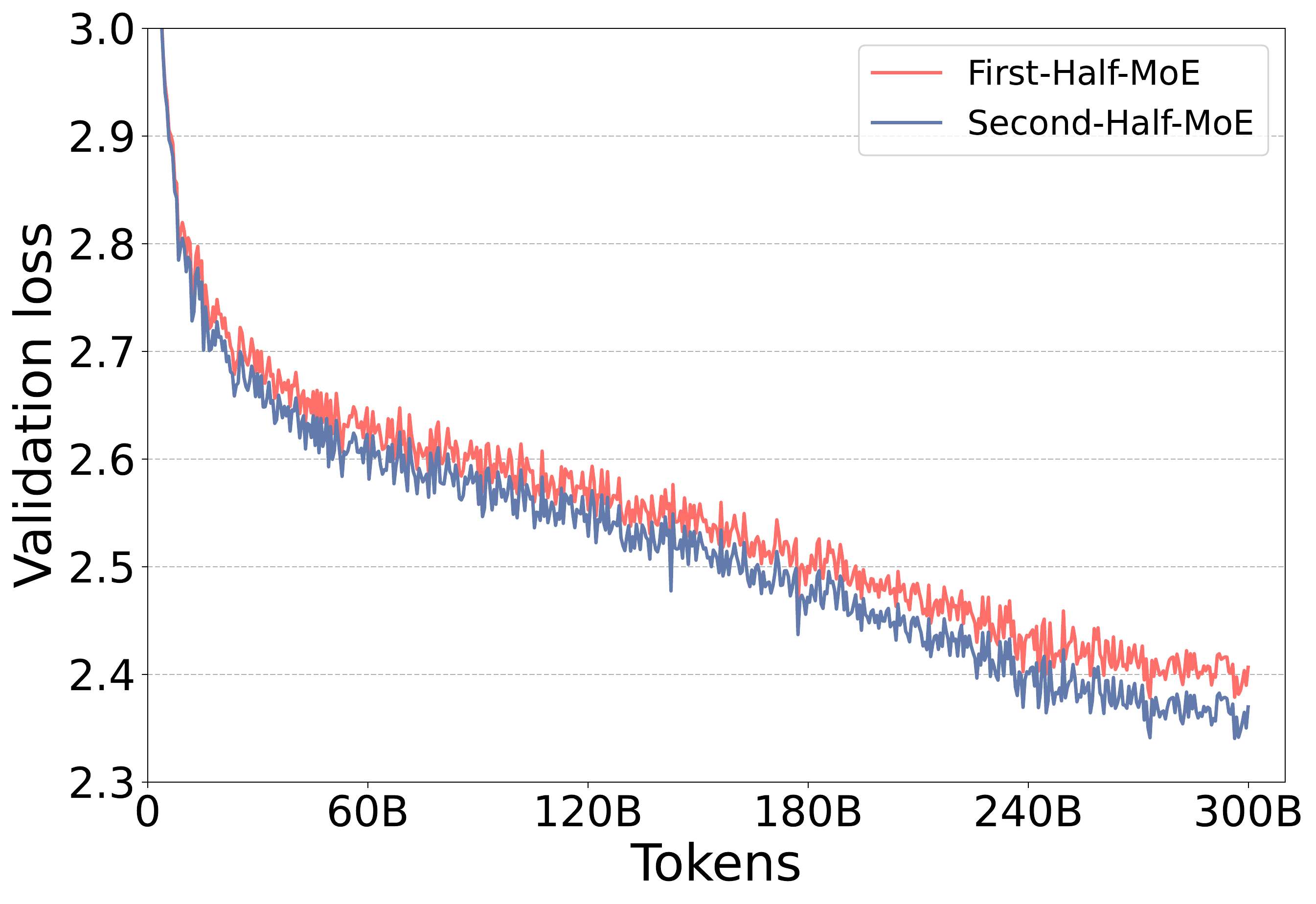}
\includegraphics[width=0.49\textwidth]{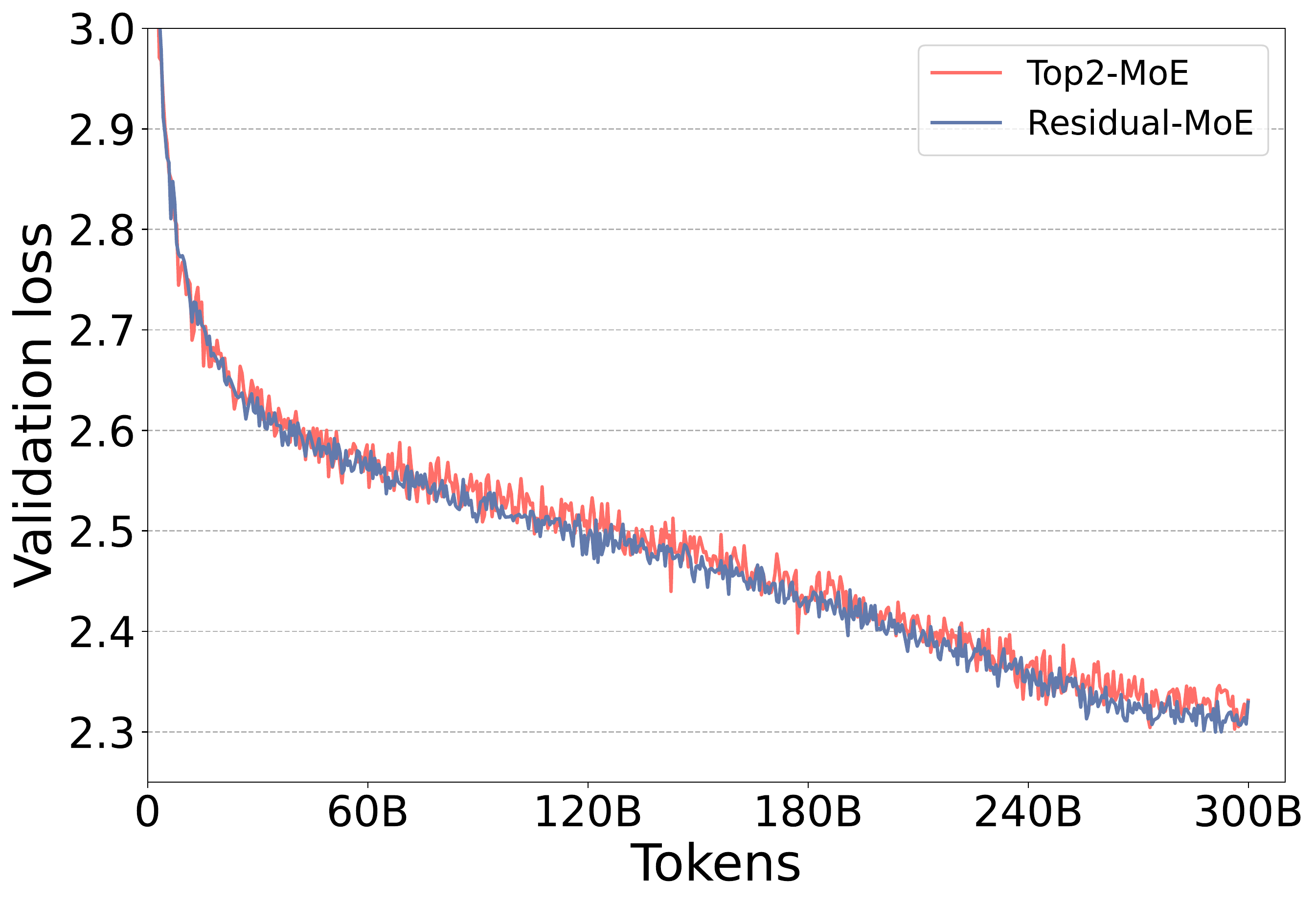}
\caption{The validation loss of First-Half-MoE/Second-Half-MoE (left) and Top2-MoE/Residual-MoE (right) based on 350M+MoE models.}
\label{fig:prmoe-inituition-valid-loss}
\end{figure}

\subsection{PR-MoE: Pyramid-Residual-MoE for Smaller Model Size and Fast Inference}
\label{sec:pr_moe}
\subsubsection{Two Observations and Intuitions}
\label{sec:two_intuitions_and_observations}
\textbf{Phenomenon-I} 
First, the standard MoE architecture has the same number and structure of experts in all MoE layers. 
This reminds us a fundamental question in machine learning community: do all the layers in a Deep Neural Network learn the same representation? 
This question has been well-studied in Computer Vision (CV): shallow layers (close to inputs) learn general representations and deep layers (close to outputs) learn more objective specific representations~\cite{zeiler2014visualizing}.
This also inspired transfer learning in CV to freeze shallow layers for finetuning~\cite{yosinski2014transferable}.
This phenomenon, however, has not been well-explored in Natural Language Processing, particularly for MoE architectures. 

To investigate this question, we compare the performance of two different Half-MoE architectures based on the 350M+MoE model. 
More specifically, a) we put MoE layers in the first half layers of the model and leave the second half of layers identical to dense model (referred to as First-Half-MoE), 
and b) we switch the MoE layers to the second half and use dense at the first half (referred to as Second-Half-MoE). 
The results are shown in~\fref{fig:prmoe-inituition-valid-loss} (left).
As can be seen, Second-Half-MoE has significantly better performance than its counterpart. 
This confirms that not all MoE layers learn the same level of representations. Deeper layers benefit more from large number of experts. For simplicity, we refer to this phenomenon as Phenomenon-I.

\textbf{Phenomenon-II}
Second, to improve the generalization performance of MoE models, there are two common methods: 
(1) increasing the number of experts while keeping the expert capacity (aka for each token, the number of experts it goes through) to be the same; 
(2) doubling the expert capacity at the expense of slightly more computation (33\%) while keeping the same number of experts. 
However, for (1), the memory requirement for training resources needs to be increased due to larger number of experts; 
for (2), higher capacity also doubles the communication volume which can significantly slow down training and inference.  
Is there a way to keep the training/inference efficiency while getting generalization performance gain?

One intuition of why larger expert capacity helps accuracy is that those extra experts can help correct the ``representation'' of the first one. 
However, does this first expert need to be changed every time? 
Or can we fix the first expert and only assign different extra experts to different tokens? 
To investigate this unknown property, we perform a comparison in two ways (1) doubling the capacity (referred to as Top2-MoE), and (2) fixing one expert and varying the second expert across different experts (referred to as Residual-MoE). 
Particularly, for (2), a token will always pass a dense MLP module and an expert from MoE module, which can be viewed as a special case of residual network. 
Afterward, we add the output of these two branches together to get the final output. 
The main intuition is to treat the expert from MoE module as an error correction term of the dense MLP module. 
Such that, we can achieve the benefit of using 2 experts per layer with the same amount communication volume as Top-1 gating function. 
We perform the comparison for the 350M+MoE model with 32 experts and the validation curves are presented in~\fref{fig:prmoe-inituition-valid-loss} (right). 
We find out that the generalization performance of these two (aka Top2-MoE and Residual-MoE) is on-par with each other. 
However, the training speed of our new design, Residual-MoE, is more than 10\% faster than Top2-MoE due to the communication volume reduction. 
This phenomenon is referred to as Phenomenon-II.

\begin{figure}[t]
\centering
\includegraphics[width=0.95\textwidth]{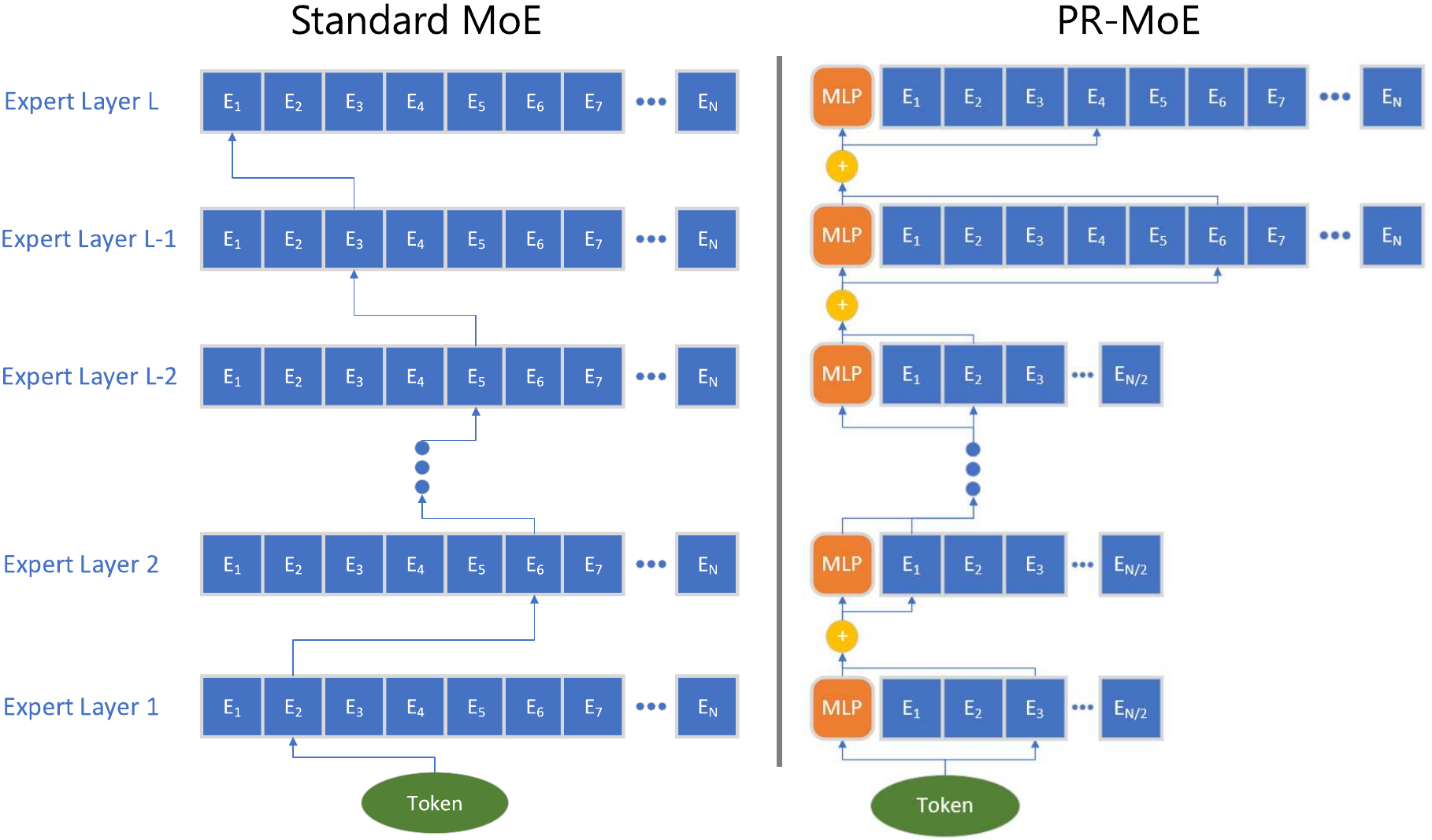}
\caption{The illustration of standard MoE (left) and \prmoe (right). 
}
\label{fig:standard_pr_moe}
\end{figure}
\subsubsection{Pyramid Residual MoE Architecture}
Based on the above, we propose our novel MoE architecture. 
As Phenomenon-I in~\sref{sec:two_intuitions_and_observations} suggested that leveraging MoE at the later layers bring more benefits, our new architecture utilizes more experts in the last few layers as compared to previous layers. 
This gives the \pyramidmoe design, where we show an example in~\fref{fig:standard_pr_moe} (right)--the last two layers have 2x experts as the previous layers.   
Meanwhile, considering Phenomenon II, we propose the \residualmoe architecture, where each token separately passes one fixed MLP module and one chosen expert as shown in \fref{fig:standard_pr_moe} (right), where orange blocks are the fixed MLP.

By combining \pyramidmoe and \residualmoe together, we have our Pyramid-Residual-MoE model (\prmoe in short), where all standard MoE layers are replaced by the new PR-MoE layer. 
\fref{fig:standard_pr_moe} shows the illustration of standard-MoE and PR-MoE architectures.

\subsubsection{System Design}
In this section, we begin by discussing how an MoE model can be trained efficiently using expert parallelism. 
Then we discuss the limitation of such an approach when applying it to \prmoe model. 
Finally, we discuss how we can extend existing expert-parallelism based training systems to efficiently train \prmoe models.

\paragraph{Efficiently Training an MoE model}
Training an MoE model efficiently requires having sufficiently large batch size for each expert in the MoE module to achieve good compute efficiency. 
This is challenging since the number of input tokens to an MoE is partitioned across all the experts which reduces the number of tokens per expert proportionally to the number of experts when compared to the rest of the model where no such partition is done. 
The simplest way to avoid this reduction in tokens per expert is to train the model with data parallelism in combination with expert parallelism \cite{deepspeed-moe-paper} equal to the number of experts. 
This increases the aggregate tokens in the batch per MoE replica that will be partitioned between the experts, resulting in no reduction of the tokens per expert compared to rest of the model. 

\paragraph{Challenges of PR-MoE}
Designing a training infrastructure that can efficiently train \prmoe model is non-trivial due to the presence of different number of experts at different stages of the model. 
As discussed above, the most efficient approach to training MoE based models is to make expert parallelism equal to the number of experts, to avoid reducing input tokens per experts. 
However, due to variation in the number of experts in \prmoe, there is no single expert parallelism degree that is optimal for all MoE layers.
Furthermore, if expert parallelism is set to the smallest number of experts in the model, then it would require multiple experts per GPU for MoE layers with larger number of experts, resulting in poor efficiency due to reduced batch size per expert, as well as an increase in memory required per GPU.
On the other hand, if we set the expert parallelism to be the largest number of experts in the model, then this would result in a load balancing problem, where some GPUs have more experts to process than the others, ultimately limiting training throughput efficiency.

\paragraph{DeepSpeed-MoE with Multi-expert and Multi-data Parallelism Support}
To address these challenges, we develop and implement a flexible multi-expert and multi-data parallelism design on top of DeepSpeed-MoE, that allows for training different parts of the model with different expert and data parallelism degree. 
For instance, a PR-MoE model running on 128 GPUs, with 32, 64, and 128 experts at different MoE layers, can be trained with 128-way data parallelism for the non-expert parallelism, and \{32, 64, 128\} expert parallelism plus \{4, 2, 1\} data parallelism for MoE parameters.
Note that each GPU can now train exactly 1 expert per MoE layer regardless of the number of experts in it, resulting in no reduction in input tokens per expert, no load-imbalance, or increase in memory requirements per GPU.

Through this flexible extension, DeepSpeed-MoE~\cite{deepspeed-moe-paper} can train \prmoe models, along with any other future MoE variations that may require different experts at different stages of the model, without compromising on training efficiency or the memory requirements.

\subsubsection{Evaluation of PR-MoE}
\label{subsec:pr-moe}

\begin{table*}[t]
\centering
  \footnotesize
  \caption{Zero-shot evaluation comparison (last six columns) between standard MoE and \prmoe.}
  \label{table:standard_prmoe_comparison}
  \begin{tabular}{rccccccc}
  \hline
  Model (num. params) & LAMBADA & PIQA & BoolQ & RACE-h & TriviaQA & WebQs\\
  \hline
  350M+MoE-128 (13B) & 62.70 & \textbf{74.59} & \textbf{60.46} & 35.60 & \textbf{16.58} & 5.17 \\
  350M+PR-MoE-32/64 (\textbf{4B}) & \textbf{63.65} & 73.99 & 59.88 & \textbf{35.69} & 16.30 & \textbf{4.73} \\
  \hline
  1.3B+MoE-128 (52B) & 69.84 & 76.71 & 64.92 & \textbf{38.09} & \textbf{31.29} & 7.19 \\
  1.3B+PR-MoE-64/128 (\textbf{31B}) & \textbf{70.60}	& \textbf{77.75} & \textbf{67.16} & \textbf{38.09} & 28.86 & \textbf{7.73} \\
  \hline
  \end{tabular}\vspace{-0.2cm}
\end{table*}

\paragraph{Comparison between \prmoe and \standardmoe}
We evaluate our new architecture, \prmoe on two different sizes of models, i.e., base size of 350M and 1.3B, and compare the performance with larger \standardmoe architectures.  
More specifically, we compare 350M+PR-MoE-32/64 with 350M+MoE-128, and we compare 1.3B+PR-MoE-64/128 with 1.3B+MoE-128.

The results are shown in~\tref{table:standard_prmoe_comparison}. 
For both 350M and 1.3B cases, our \prmoe uses much fewer parameters but achieves comparable accuracy as \standardmoe models. 
Particularly, (1) for 350M case, \prmoe only uses less than 1/3 of the parameters as \standardmoe; 
(2) for 1.3B case, \prmoe only uses about 60\% of the parameters as \standardmoe, while achieving similar accuracy.

\begin{figure}[t]
\centering
\includegraphics[width=0.5\textwidth]{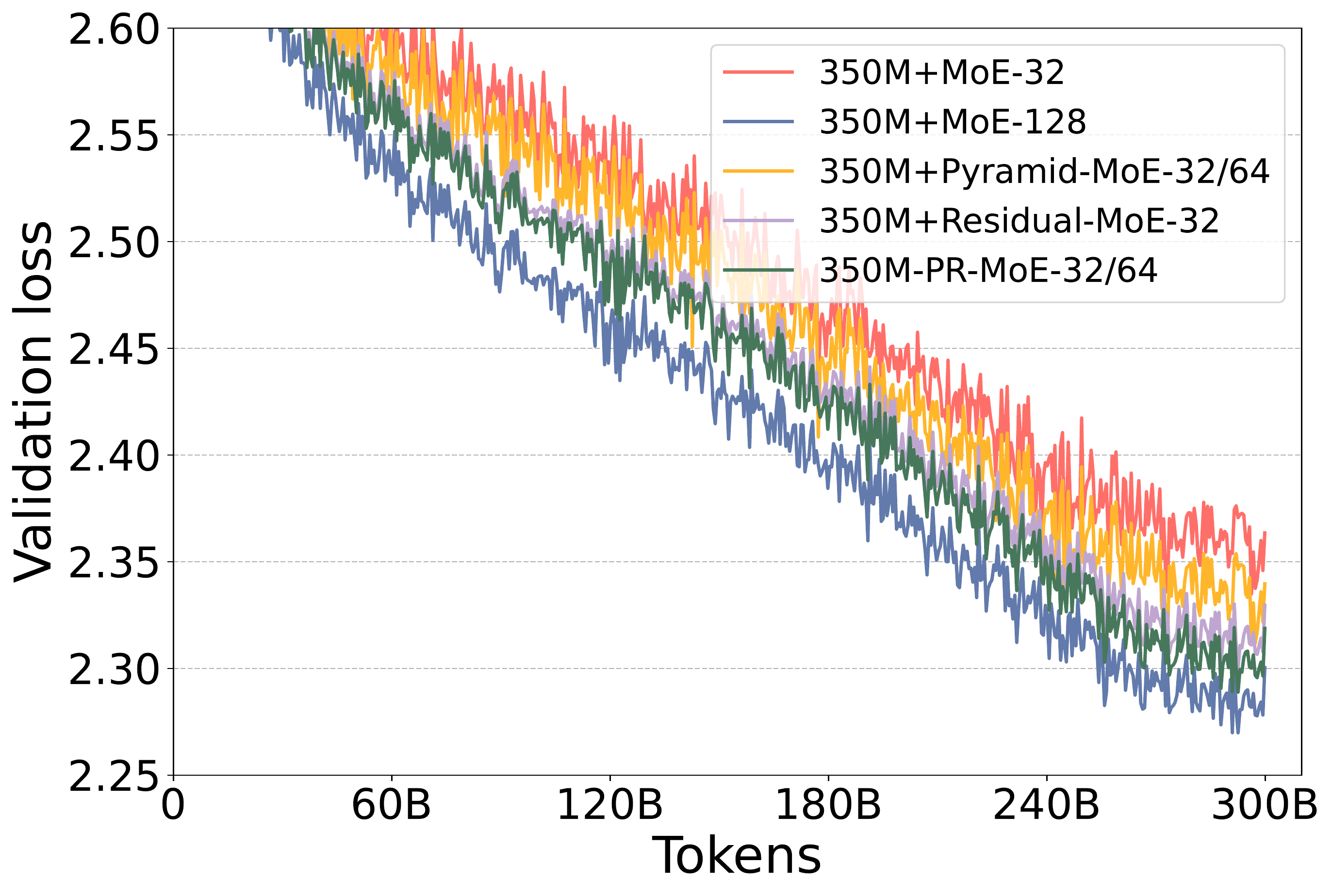}
\caption{The validation curves of different MoE models based on 350M+MoE.}
\label{fig:validation_all_moe}
\end{figure}

\paragraph{Ablation Study of Different MoE Architectures}
To fully study the performance of different MoE architectures, particularly the comparison between \standardmoe and \residualmoe/\pyramidmoe/\prmoe, 
we evaluate 5 different MoE-based models, including 350M+MoE-32, 350M+MoE-128, 350M+\pyramidmoe-32/64 (which has 10 MoE layers using 32 experts and 2 MoE layers using 64 experts), 350M+\residualmoe-32, and 350M+\prmoe-32/64 (same expert setting as 350M+\pyramidmoe-32/64).

The validation curves for the full training are shown in~\fref{fig:validation_all_moe}.
As can be seen, the validation loss gap between 350M+MoE-128 and 350M+MoE-32 can be significantly reduced by \pyramidmoe and \residualmoe. 
When we use \prmoe, a combination of \pyramidmoe and \residualmoe, the loss gap can be further reduced to around 0.01, demonstrating \prmoe's great parameter efficiency with minimum quality impact.  

\input{_s4_2_compression}

%% file: _s4_2_compression.tex
\subsection{Mixture-of-Students: Distillation for Even Smaller Model Size and Faster Inference }
\label{sec:mos}

Model compression and knowledge distillation present additional opportunities to improve inference performance further.  While they are many ways for model compression, such as quantization~\cite{q-bert,hawq,hawq-v2} and pruning~\cite{sparse-attention,block-pruning-transformer}, our current efforts focus on layer reduction through knowledge distillation~\cite{kd-hinton} (KD) – reducing both model size and model computation, and preserving MoE structure at student model.  

KD has been proven to be a successful way to compress a large model into a small one, which contains much fewer parameters and computations but still obtaining competitive results. There have been some works that apply KD to task-specific distillation of pre-trained large LMs into small models~\cite{distill-bert,patient-kd,mini-lm,mobile-bert}. However, they only consider small transformers (a few hundreds of parameters) and dense encoder-based LM models (e.g., BERT).
In contrast, we focus on studying KD for sparse MoE-based auto-generative LMs models on multi-billion parameter scale. The only other analysis of MoE distillation we are aware of are by \cite{switch-transformer,artetxe2021efficient}, who study distillation of MoE into dense models. However, by doing so the distilled student model loses the sparse fine-tuning and inference benefits provided by MoE. In contrast, our study show that it is possible to reach similar performance, such as zero-shot evaluation on many downstream tasks, for smaller MoE model pretrained with knowledge distillation, resulting in models that are lighter and faster during inference time. 

\subsubsection{Mixture-of-Students via Staged KD}

\paragraph{Architecture Choice and Optimization Objective} 
To apply knowledge distillation for MoE, we first train a teacher MoE model. We reduce the depth of each expert branch in the teacher model to obtain a corresponding student. By doing so, the final student model that has the same sparsely gated architecture as the teacher MoE except that each expert branch has a smaller depth. For this reason, we call the resulting model Mixture-of-Students (MoS). Since MoE structure brings significant benefits by enabling sparse training and inference, our task-agnostic distilled Mixture-of-Students inherits these benefits while preserving the inference advantage over its quality equivalent dense model. After creating the MoS, we force the MoS to imitate the outputs from the teacher MoE on the training dataset. We take a general formulation of the KD loss~\cite{kd-loss} as:
\begin{equation}
     \min_{\theta} \mathbb{E}_{(x,y)\sim D}[\mathcal{L}(x;\theta) + \alpha \mathcal{L}_{KD}(x';\theta)], \label{eqn:objective} 
\end{equation}

where as a weight sum of the cross-entropy loss between predictions and the given hard label and the KL divergence loss between the predictions and the teacher’s soft label. Furthermore, given the excellent performance of PR-MoE, we combine PR-MoE together with KD (PR-MoS) to further reduce the MoE model sizes. In other words, we choose both the teacher model and the student model to be PR-MoE.

\begin{figure}[t]
\centering
\includegraphics[width=0.5\textwidth]{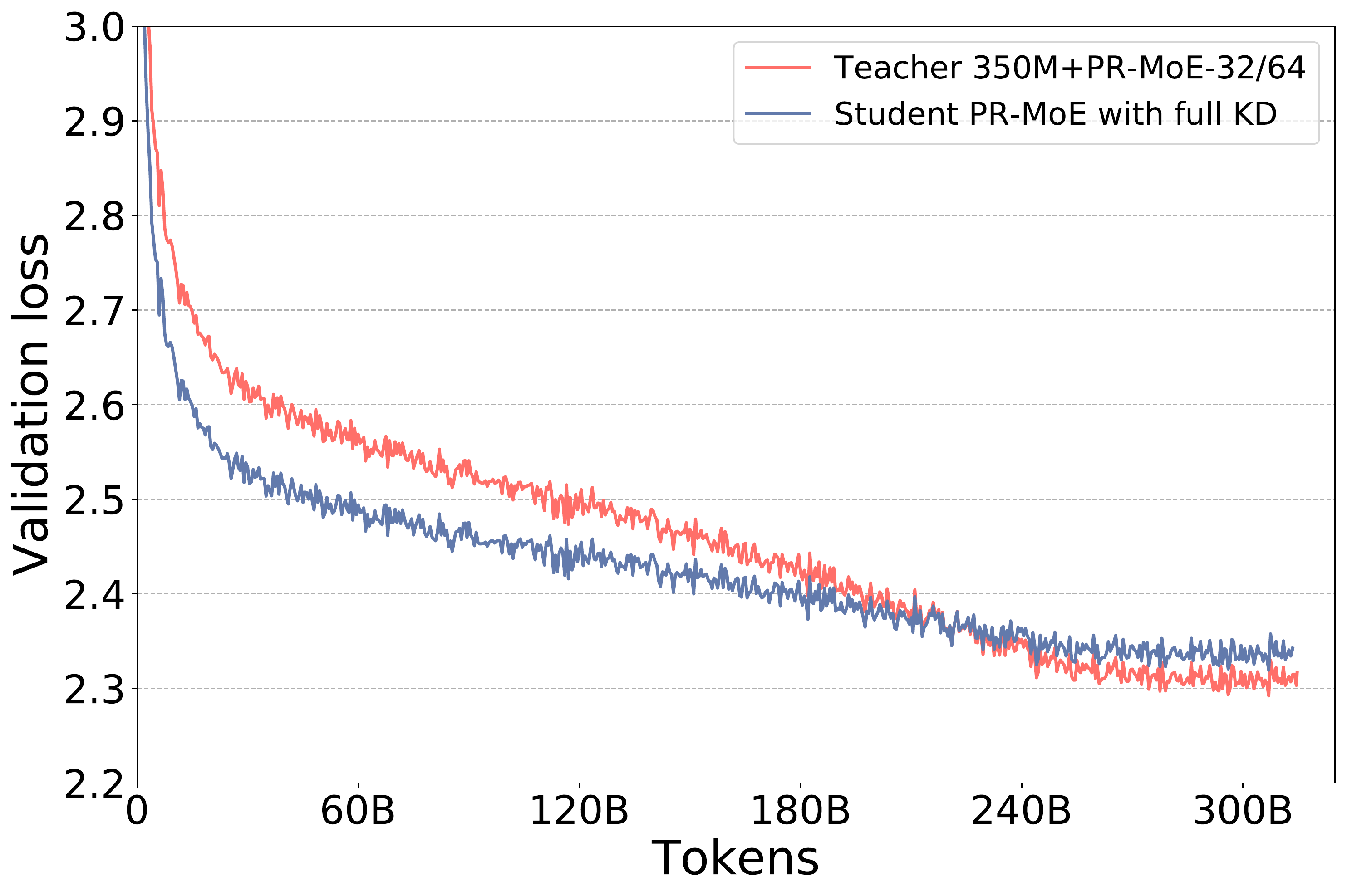}
\caption{The validation curves of training without distillation from scratch vs. performing knowledge distillation for the entire pre-training process. In the figure the student PR-MoE is trained with 21-layer. KD helps initially but starts to hurt accuracy towards the end of training.}
\label{fig:validation_full_kd}
\end{figure}

\paragraph{Improving Student Accuracy with Staged Knowledge Distillation}
One interesting observation during distilling MoE model is that using the teacher PR-MoE leads to lower student accuracy than a PR-MoE student trained from scratch (row 2 and 4 in Table~\ref{table:kd_prmoe_comparison}). As KD often improves the student generalization, this raises a question on why KD does not improve accuracy for pre-training MoE on generative language model. Since no prior experiments reported distillation experiment results on distilled MoE, we dig deeper into the results. Figure~\ref{fig:validation_full_kd} shows a comparison of validation loss between a PR-MoE trained from scratch and using knowledge distillation with its teacher. We find that while KD loss improves validation accuracy initially, it begins to hurt accuracy towards the end of training (e.g., after 400K steps).

We hypothesize that because the PR-MoE already reduces the capacity compared with the standard MoE by exploiting the architecture change (e.g., reducing experts in lower layers), further reducing the depth of the model causes the student to have insufficient capacity, making it fall into the underfitting regime. Therefore, the student PR-MoE may not have enough capacity to minimize both the training loss and the knowledge distillation loss, and might end up minimizing one loss (KD loss) at the expense of the other (cross entropy loss), especially towards the end of training.
The aforementioned hypothesis suggests that we might want to either gradually decay the impact from KD or stop KD early in the training process and perform optimization only against the standard language modeling loss for the rest of the training. 

\subsubsection{Evaluation of Mixture-of-Students}

We evaluate our approach on two different PR-MoE model configs, 350M+PR-MoE-32/64 and 1.3B+PR-MoE-64/128. We build student models by reducing the depth of the teachers to 21 (12.5\%) in both cases and compare the resulting MoS model with its teacher using the method described in the previous section. 

\begin{figure}[!ht]
\centering
\includegraphics[width=0.49\textwidth]{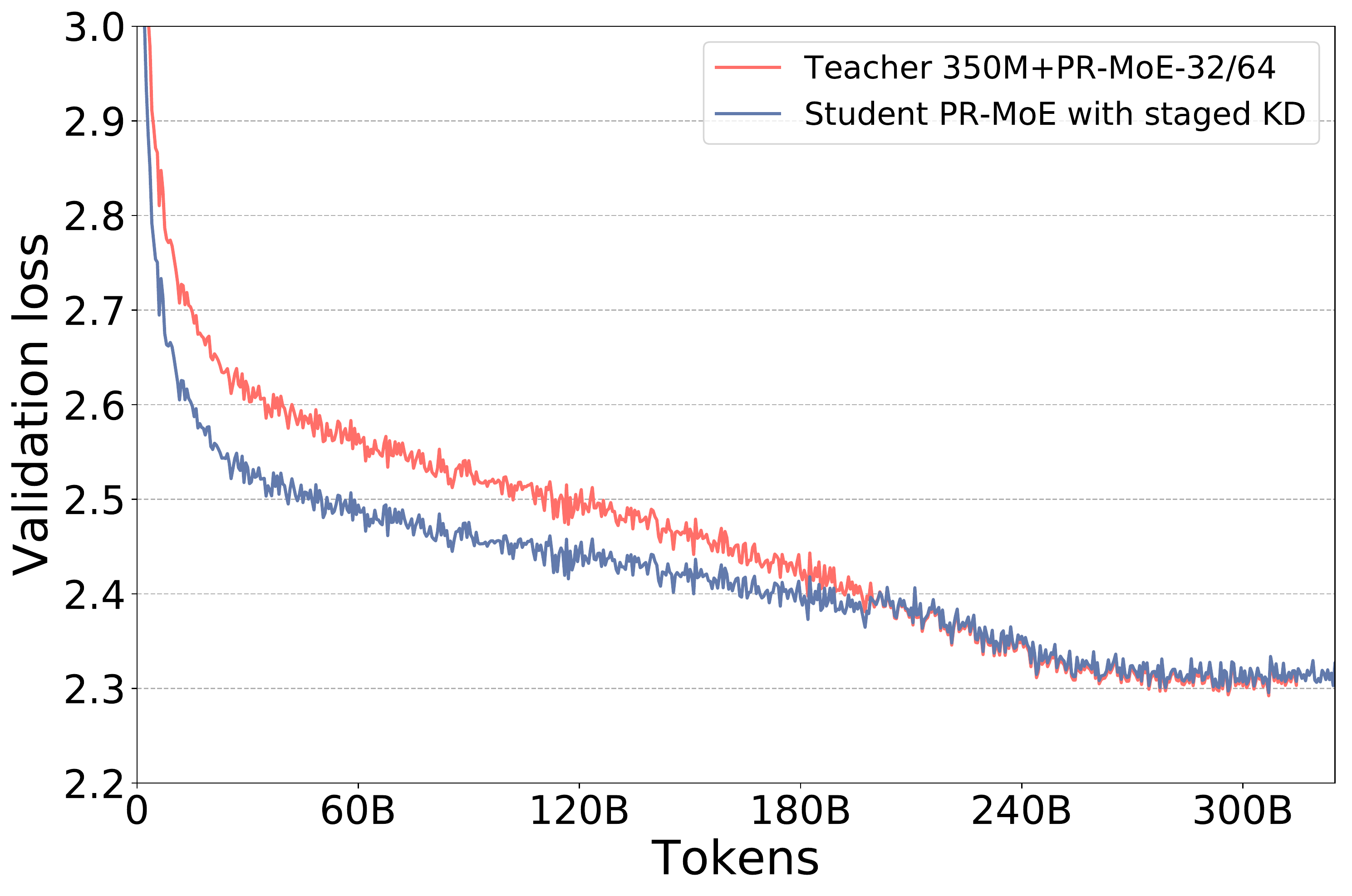}
\includegraphics[width=0.49\textwidth]{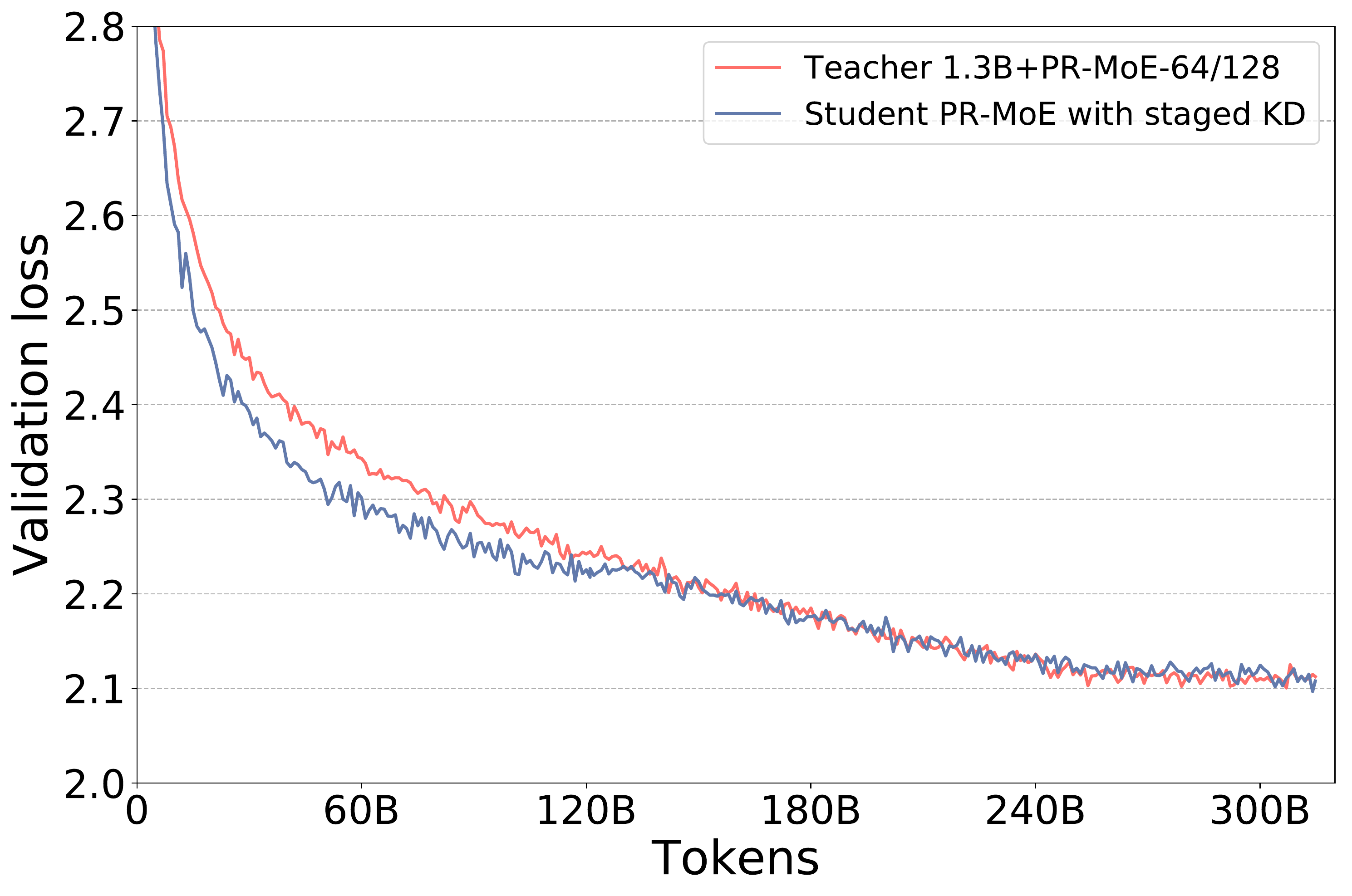}
\caption{The validation curves of training the student PR-MoE with staged knowledge distillation obtains almost the same validation loss as the teacher PR-MoE on both 350M+PR-MoE and 1.3B+PR-MoE.}
\label{fig:validation_full_kd_staged}
\end{figure}

We first evaluate how the proposed stage-KD affects the pre-training convergence. Figure~\ref{fig:validation_full_kd_staged} shows the validation curve comparison of stopping KD at 400K steps and its comparison with the teacher model. We find that this staged version of KD now gives us the promised benefit of knowledge distillation: the student model now has a similar validation curve as the teacher. The evaluation results on downstream tasks also show that the staged KD achieves much better zero-shot evaluation accuracy than applying KD for the entire training process, as shown next.

Next we perform zero-shot evaluation on a set of NLP tasks. The results on each of the tasks are shown in Table~\ref{table:kd_prmoe_comparison}. We make a few observations. First, with the same amount of depth reduction but without MoS-based KD (row 2), the PR-MoE model encounters noticeable accuracy drop on several tasks such as LAMBADA (1.3 points) and BoolQ (7.5 points), indicating that directly reducing the expert depth can hurt the model accuracy. Second, with staged KD (row 4), we are able to improve the student PR-MoE's performance and observe accuracy improvements on 5 out of 6 tasks. 
Notably,  1.1 points improvement for LAMBADA, 6.5 points higher for BoolQ, 1.7 points higher for RACE-h, 4.5 points higher for TriviaQA.  One exception is PIQA, in which case the student PR-MoE experiences some small accuracy drop. These results indicate the effectiveness of our proposed Mixture-of-Students method as a novel KD technique for MoE models. Third, performing KD for the entire training process (full KD, row 3) hurts the downstream task accuracy on LAMBADA (0.8 points lower) and PIQA (0.7 points lower). As explained in the previous part, this is because the student model does not have sufficient capacity to optimize both the KD loss and the standard LM loss towards the end of training, due to under-fitting. In contrast, our proposed staged-KD is able to resolve this issue and brings the promised benefits from KD. Overall, the distilled MoE model through staged KD achieves an average accuracy of 42.87 and 47.96, retaining 99.5\% and 99.1\% performance of the 350M (43.08) and 1.3B teacher model (48.37) despite having 12.5\% fewer layers. This enables an additional latency reduction and throughput reduction for inference, which we show in the next section.

\begin{table*}[!ht]
\centering
  \footnotesize
  \caption{Zero-shot evaluation comparison (last six columns) between PR-MoE and PR-MoE + MoS.}
  \label{table:kd_prmoe_comparison}
  \begin{tabular}{rccccccc}
  \hline
  Model (num. params) & LAMBADA & PIQA & BoolQ & RACE-h & TriviaQA & WebQs\\
  \hline
  350M+PR-MoE+L24({4B}) & {63.65} & 73.99 & 59.88 & {35.69} & 16.30 & {4.73} \\
  350M+PR-MoE+L21 ({3.5B}) & {62.33} & 73.88 & 52.35 & {32.54} & 8.81 & {4.48} \\
  350M+PR-MoE+L21+KD only ({3.5B}) & {61.56} & 73.18 & 57.89 & {33.78} & 12.13 & {4.87} \\
  350M+PR-MoE+L21+MoS ({3.5B}) & 63.46 & 73.34 & 58.07 & 34.83 & {13.69} & 5.22 \\
  \hline
  1.3B+PR-MoE+L24 ({31B}) & {70.60}	& {77.75} & {67.16} & {38.09} & 28.86 & {7.73} \\
  1.3B+PR-MoE+L21+KD only ({27B}) & {69.73}	& {76.93} & {64.16} & {36.17} & 26.17 & {6.25} \\
  1.3B+PR-MoE+L21+MoS ({27B}) & 70.17 & 77.69 & 65.66 & 36.94 & 29.05 & 8.22 \\
      \hline
  \end{tabular}
\end{table*}

As the conclusion of Section~\ref{sec:moe_parameter_efficiency}, during training PR-MoE leads to up to 3x drop in memory consumption compared to the original standard MoE model. During inference, PR-MoE and MoS together reduces the MoE model size by up to 3.7x while retaining the model accuracy. This significantly benefits inference in latency and cost, as we describe in the next section.

%% file: _s5_1_inference-system.tex
\section{DeepSpeed-MoE Inference: Serving MoE Models at Unprecedented Scale and Speed}
\label{sec:optimizing_moe_inference_latency}

Optimizing inference latency and cost is crucial for MoE models to be useful in practice. 
During inference, the batch size is generally small, so the inference latency of an MoE model depends primarily on the time it takes to load the model parameters from the main memory, contrasting with the conventional belief that lesser compute should lead to faster inference. 
Therefore, the MoE inference performance depends on two main factors: the overall model size and the overall achievable memory bandwidth.

In the previous section, we presented PR-MoE and MoS to reduce the MoE model size while preserving the model accuracy.
This section presents our system optimization solutions to maximize the achievable memory bandwidth by creating a multi-GPU MoE inference system that leverages the aggregated memory bandwidth across dozens of distributed GPUs to speed up inference. 
Together, \textbf{DeepSpeed-MoE} (\textbf{DS-MoE} in short) offers an unprecedented scale and efficiency to serve massive MoE models with $7.3$x better latency and lower cost compared to baseline MoE systems, and up to $4.5$x faster and $9$x cheaper MoE inference compared to quality-equivalent dense models. DS-MoE is part of a larger DeepSpeed-inference effort presented in~\cite{ds-inference-paper}.

\subsection{Design of DeepSpeed-MoE Inference System} 
\textit{MoE inference performance is an interesting paradox:} 
\begin{itemize}
    \item From the best-case view, each input token of an MoE model (with top-1 gating) only activates a single expert at each MoE layer, resulting in a critical data path that is equivalent to the base dense model size, orders-of-magnitude smaller than the actual model size. 
    For example, when inferencing a 1.3B+MoE-128 model, each input token needs just 1.3 billion parameters, even though the overall parameter size is 52 billion. 
    \item From the worst-case view, the aggregate parameters needed to process a batch of tokens (e.g., a sentence or a paragraph of text) can be as large as the full model size (since different tokens could activate different experts), the entire 52 billion parameters in the previous example, making it challenging to achieve short latency and high throughput.
\end{itemize}

The design goal of DeepSpeed-MoE inference system is to steer the performance toward the best-case view. 
It is achieved through three sets of well-coordinated optimizations:
\begin{itemize}
    \item Carefully partition the model and embrace different types of parallelism; group and route all tokens with the same critical data path together to reduce data access per device and achieve maximum aggregate bandwidth;
    \item Optimize communication scheduling with parallelism coordination to effectively group and route tokens; and 
    \item Optimize transformer and MoE related kernels to improve per-device performance.
\end{itemize}
We present an in-depth discussion of these optimizations in the next three sections.   

\subsection{Flexible Combination of Tensor-Slicing, Expert-Slicing, Data Parallelism, and Expert Parallelism}

\begin{figure}[htbp]
\centering
\includegraphics[width=\textwidth]{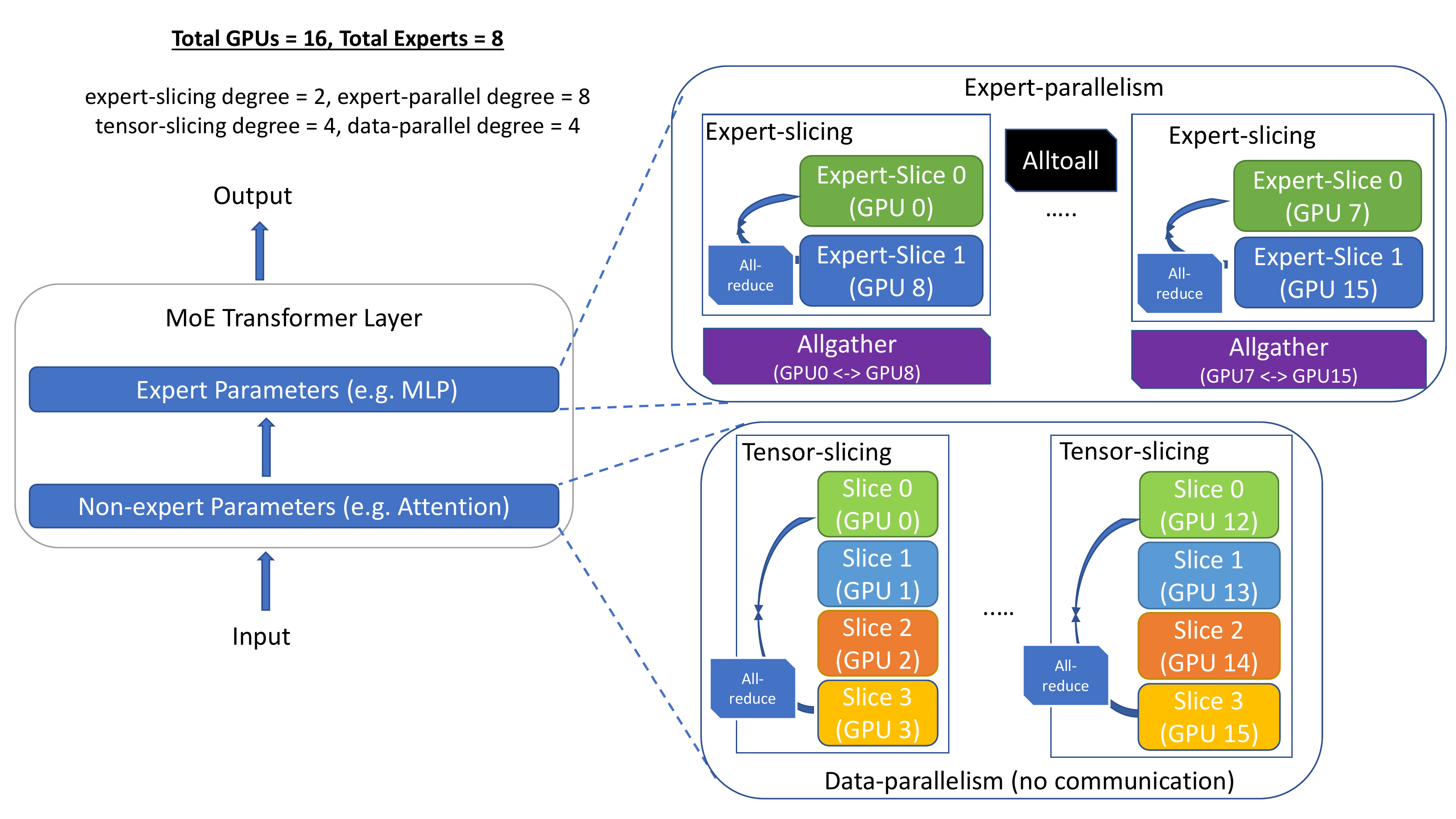}
\caption{DS-MoE design that embraces the complexity of multi-dimensional parallelism for different partitions (expert and non-expert) of the model.}
\label{fig:inference-design}
\end{figure}

To achieve low latency and high throughput at an unprecedented scale for MoE, we design our inference system to minimize the critical data path per device, maximize the achievable aggregate memory bandwidth, and offer ample aggregate memory simultaneously to enable massive model sizes by using (1) expert parallelism~\cite{deepspeed-moe-paper} and slicing on expert parameters and (2) data parallelism and tensor-slicing for non-expert parameters.
\fref{fig:inference-design} illustrates a single MoE Transformer layer, which contains both experts (e.g., MLP) and non-expert parameters (e.g., Attention), and how we use a combination of parallelism strategies to process each component. Below we describe how the model and data are partitioned, and the details of each form of parallelism we use to deal with each piece.

\paragraph{Expert Parallelism and Expert-slicing for Expert Parameters}
As illustrated in the MoE performance paradox, while each token only activates a single expert at each MoE layer, for batch inference with multiple tokens, the aggregate parameters needed for all the tokens can be as large as the entire set of parameters, making it challenging to achieve both low latency and high throughput. To address this issue, we partition experts across devices, group all input tokens assigned to the same experts under the same critical data path, and parallelize processing of the token groups with different critical paths among different devices using expert parallelism. 

In the example of 1.3B+MoE-128, when expert parallelism is equal to 128, each GPU only processes a single token group corresponding to the experts on that device. 
This results in a sequential path that is 1.3 billion parameters per device, 5x smaller than its quality-equivalent dense model with 6.7B parameters. 
Therefore, in theory, an MoE-based model has the potential to run up to 5x faster than its quality-equivalent dense model using expert parallelism assuming no communication overhead, a topic we discuss in the next section. 

In addition, we propose “expert-slicing” to leverage the concept of tensor-slicing for the parameters within an expert, which partitions the expert parameters horizontally/vertically across multiple GPUs.
This additional dimension of parallelism is helpful for latency stringent scenarios that we scale to more devices than the number of experts.

\paragraph{Data Parallelism and Tensor-slicing for Non-expert Parameters} 
While expert parallelism reduces the number of expert parameters in the critical path per device, it does not reduce the non-expert parameters in the critical path. 
This leads to two limitations: (1) the maximum size of non-expert parameters in the MoE model that can be inferenced is limited by single device memory, and (2) the execution latency of the non-expert components of the model is limited by single device memory bandwidth. 

We use tensor-slicing within a node to address these bottlenecks, allowing for hundreds of billions of non-expert parameters by leveraging aggregate GPU memory, while also leveraging the aggregate GPU memory bandwidth across all GPUs within a node. 
While it is possible to perform tensor-slicing across nodes, the communication overhead of tensor-slicing along with reduced compute granularity generally makes inter-node tensor-slicing infeasible. 
To scale non-expert parameters across multiple nodes, we use data-parallelism by creating non-expert parameter replicas processing different batches across nodes which incur no communication overhead or reduction in compute granularity.

\paragraph{Synergy of Multidimensional Parallelism} 
By combining expert-parallelism and expert-slicing with tensor-slicing and data-parallelism, DS-MoE inference can scale a multi-trillion parameter MoE model (with trillions of expert parameters and hundreds of billions of non-expert parameters) to dozens or even hundreds of devices across nodes. 
The aggregate bandwidth across these devices and minimized critical data path per device open up the opportunity to enable low latency and high throughput inference at an unprecedented scale. 
However, getting there still requires high performance communication collectives and single device kernels, which we talk about next.

\subsection{Optimized Communication Subsystem: Grouping and Routing Tokens More Efficiently} \label{sec:opt-alltoall}
Expert parallelism requires all-to-all communication between all expert parallel devices.
By default, DS-MoE uses NCCL for this communication via ``torch.distributed'' interface, but we observe major overhead when it is used at scale (more results in Section~\ref{subsec:moe-perf}). 
To optimize this, we develop a custom communication interface to use Microsoft SCCL~\cite{sccl-paper} and achieve better performance than NCCL. 
Despite the plug-in optimizations, it is difficult to scale expert parallelism to many devices as the latency increases linearly with the increase in devices. 
To address this critical scaling challenge, we design two new communication optimization strategies that exploit the underlying point-to-point NCCL operations and custom CUDA kernels to perform necessary data-layout transformations.

\begin{figure}[htbp]
\centering
\includegraphics[width=0.9\textwidth]{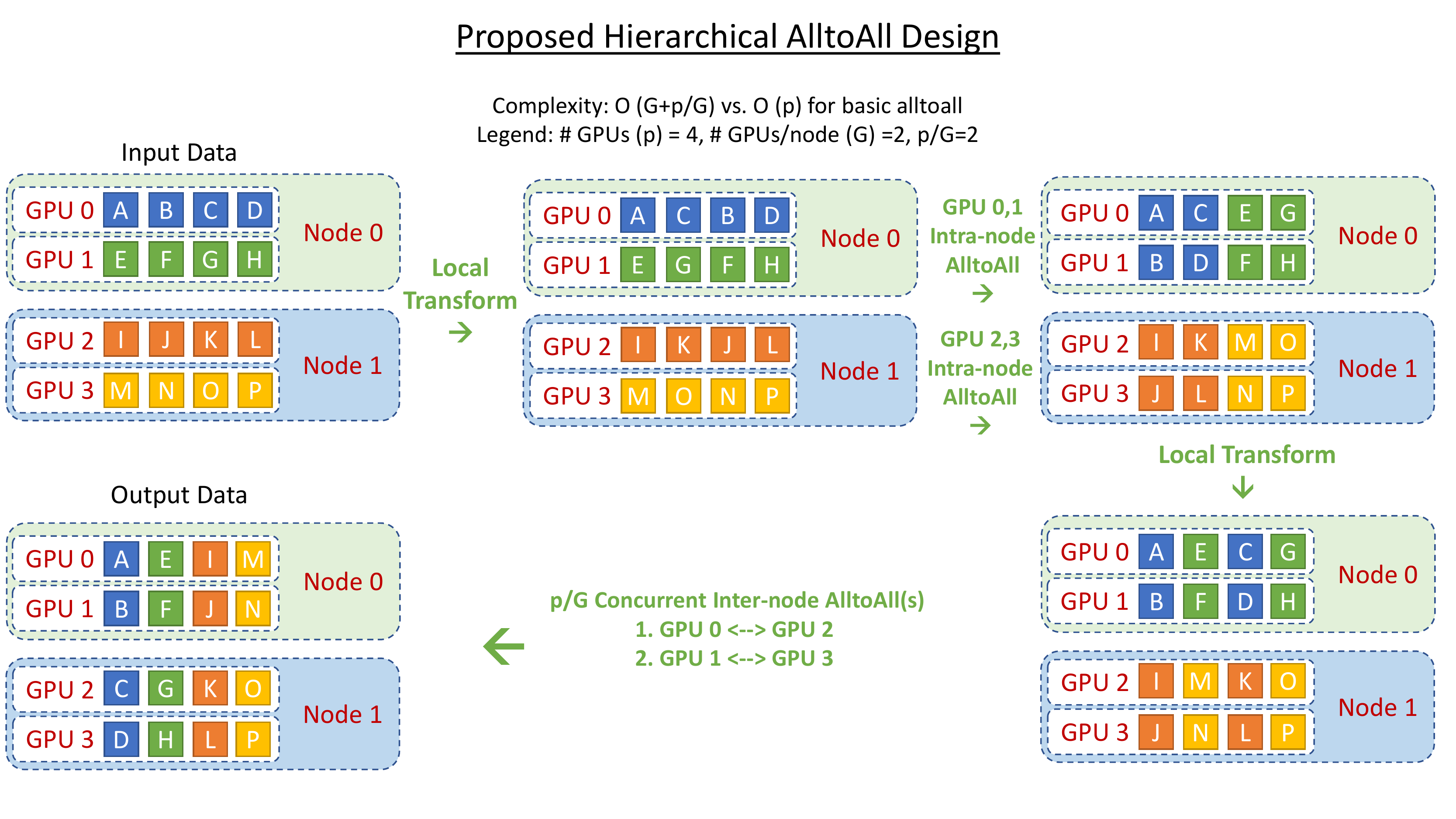}
\caption{Illustration of the proposed hierarchical all-to-all design.}
\label{fig:halltoall}
\end{figure}

\paragraph{Hierarchical All-to-all} 
Hierarchical tree-based algorithms are often used with communication collectives like allreduce, broadcast, etc to reduce the number of communication hops.
We implement a hierarchical all-to-all as a two-step process with a data-layout transformation, followed by an intra-node all-to-all, followed by a second data-layout transformation, and a final inter-node all-to-all. 
This reduces the communication hops from $O(p)$ to $O(G+p/G)$, where $G$ is the number of GPUs in a node and $p$ is the total number of GPU devices. 
\fref{fig:halltoall} shows the design overview of this implementation. 
Despite the $2$x increase in communication volume, this hierarchical implementation allows for better scaling for small batch sizes as communication at this message size is more latency-bound than bandwidth-bound. 

\paragraph{Parallelism Coordinated Communication Optimization} Combining expert parallelism and tensor-slicing with data parallelism within a single model is non-trivial in terms of handling communication effectively. 
Tensor-slicing splits individual operators across GPUs and requires all-reduce between them, while expert parallelism places expert operators across GPUs without splitting them and requires all-to-all between them. 
A naïve approach to handle these communications is to treat each parallelism as a black box, performing the required communication independently. 
However, this would lead to sub-optimal performance. 

\begin{figure}[htbp]
\centering
\includegraphics[width=0.9\textwidth]{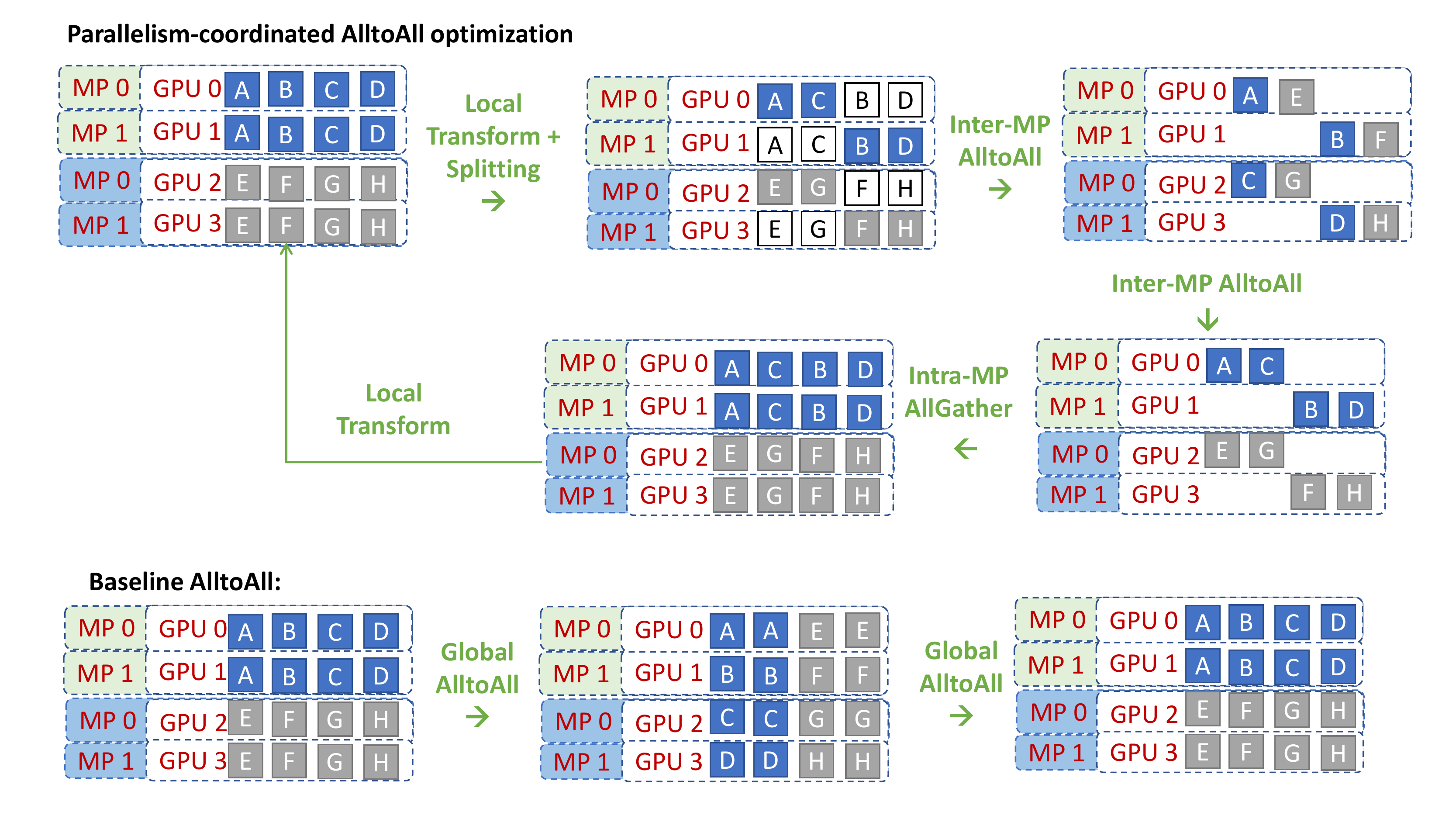}
\caption{Illustration of the parallelism coordinated communication.}
\label{fig:palltoall}
\end{figure}

The all-reduce operation in tensor-slicing replicates data among the involved devices. When executing tensor parallel operators followed by expert parallel operators, this replication allows creating an optimized communication schedule for the all-to-all operator that does not require communicating between all the expert parallel processes. 
Instead, the all-to-all can happen within just the subset of devices that share the same tensor-slicing rank, since the data across tensor parallel ranks are replicated (\fref{fig:palltoall}). 
As a result, the latency of all-to-all is bounded by $O(p/L)$ instead of $O(p)$ where $L$ is the tensor-slicing parallelism degree and $p$ is the total number of GPU devices.

Similarly, when executing expert parallel operator followed by tensor-slicing operators, the final all-to-all can be done in the same way, but this time followed by an allgather operator between tensor parallel ranks to replicate the data needed by tensor-slicing (\fref{fig:palltoall}). 
This reduces the latency overhead from $O(p)$ to $O(p/L) + O(L)$. 

This reduced latency overhead allows better scaling to a large number of devices. 
For example, when scaling to $128$ GPUs with $8$-way tensor-slicing and $128$-way expert parallelism, this approach reduces the latency overhead of the all-to-all from $(128C_1 + C_2)$ to $(16C_1+C_2)$ due to 8-way tensor-slicing, where $C_1$ and $C_2$ are some constants determined by point-to-point latency, message size, and bandwidth.

\subsection{Highly Optimized Transformer and MoE Related Kernels}
\label{sec:opt-kernels}

DS-MoE inference system consists of highly optimized multi-GPU transformer kernels as well as highly optimized MoE related kernels. 
We use DeepSpeed inference kernels for maximizing bandwidth utilization for the non-expert transformer layers. 
Please see \cite{ds-inference-blog} to learn more.

In this paper, we focus on the MoE related operators for performing gating and the different data layout transformations conventionally implemented using sparse-dense einsums in literature. 
At a high level, we optimize these operators by implementing them as explicit data layout transformations instead of highly sparse-dense einsums, to reduce the compute complexity from cubic to quadratic. 
We also fuse most of these operators into a single kernel. 

More specifically, the MoE-related computation consists of three major components: 
\begin{itemize}
    \item The gating function that determines the assignment of tokens to experts, where the result is represented as a sparse tensor (a one-hot vector representing the assigned expert for each token in the sequence).
    \item A sparse einsum operator, between the one-hot tensor and all the tokens, that sorts the ordering of the tokens based on the assigned expert id.
    \item A final einsum at the end of the MoE computation that scales and re-sorts the tokens back to their original ordering.
\end{itemize}	

The sparse tensor representation in the gating function and sparse einsum operators introduces a significant latency overhead. 
First, the gating function includes numerous operations to create token-masks, select top-k experts, and perform cumulative-sum (\textit{cumsum}) to find the token-id going to each expert and sparse matrix-multiply, all of which are not only wasteful due to the sparse tenor representation, but also extremely slow due to many kernel call invocations.
Moreover, the sparse einsums have a complexity of $S\times E \times M \times c_e$, where $S$ represents the total number of tokens, $E$ represents the number of experts, $M$ represents model hidden dimension, and $c_e$ represents expert capacity ($S$, $E$, and $M$ are the main complexity factors, while $c_e$ is normally very small). In this equation, $(E-1)$ out of $E$ operators for each token are multiplications and additions with zeros, since only one expert is typically selected to process $c_e$ tokens. This comes from the fact that generalizing the gating operations results in the einsums over several masking matrices or one-hot vectors that produce a lot of non-necessary computation with zeros to select the correct token for each expert.

We optimize these operators using dense representation and kernel-fusion.

First, we fuse the gating function into a single kernel, and use a dense token-to-expert mapping table to represent the assignment from tokens to experts, greatly reducing the kernel launch overhead, as well as memory and compute overhead from the sparse representation. 
More specifically, gating kernel includes top-k, \textit{cumsum}, and scatter operations in order to distribute the right tokens to each expert. 
The top-k operator selects the \textit{k} experts with the k-highest logits for each input token, and since \textit{k} is normally small (e.g., 1 or 2) for the MoE models, we store the best expert-indices in a mapping table rather than creating a mask for the rest of gating function operations. 
\textit{Cumsum} calculates the ID for the tokens processed by each expert, that is defined by the capacity-factor in the MoE configuration. 
We use the so-called Blelloch scan algorithm to parallelize \textit{cumsum} on GPU architecture. 
Finally, we use the mapping table and token IDs in order to route the correct tokens to the MoE experts.

Second, to optimize the remaining two sparse einsums, we implement them as data-layout transformations using the above-mentioned mapping table, to first sort them based on the expert id and then back to its original ordering without requiring any sparse einsum, reducing the complexity of these operations from $S\times E\times M\times c_e$ to $S\times M \times c_e$. 
Together with the data transformation, we use the corresponding gating logits (in the probability domain) to update the expert output. 

Combined, these optimizations result in over 6x reduction in MoE Kernel related latency.

\subsection{Performance Evaluation of DS-MoE Inference}
\label{subsec:moe-perf}

In modern production environments, powerful DL models are often served using hundreds of GPU devices to meet the traffic demand and deliver low latency. In this section, we explore how these two broad goals of \textit{high throughput} and \textit{low latency} can be realized for MoE model inference at scale. We also explore how MoE model inference is different compared to their dense counterparts.

The key questions we try to explore include:

1) What are the unique properties of MoE inference?

2) How does it perform and scale with increased model size and resources? 

3) What kind of benefits do model optimizations like PR-MoE and MoS bring for MoE model inference? 

4) How do MoE models compare to their quality-equivalent dense models with respect to latency and cost?

\noindent Using various model configurations shown in \tref{t:inference-table}, we try to answer these questions at scale (up to 256 A100 GPUs on Azure).

\begin{table}[htbp]
\caption{The configuration of different MoE models used for the performance evaluation of DS-MoE inference system. These configurations represent the standard MoE architecture cases, and we also test the case of PR-MoE and PR-MoE+MoS, which will have smaller sizes but same (projected) quality.}
\small
\label{t:inference-table}
\centering
\begin{tabular}{rcccccc}
\hline
 {Model} &  {Size (billions)}&  {\#Layers}&  {Hidden size}&  {MP degree} & {EP degree}\\
\hline
 {1.3B+MoE-128} &  52& 24& 2048 & 1 & 128\\
 \hline
 {2.4B+MoE-128} &  107.7& 16& 3584 & 1 & 128\\
\hline
 {8B+MoE-128} &  349.0&  30&  4096  &  4&  128\\
\hline
 {24B+MoE-128} &  1064.9&  40&  8192&  8 &  128\\
\hline
 {47B+MoE-128} &  2024.0&  58&  8192&  8 &  128\\
\hline
\end{tabular}
\vskip -1.5em
\end{table}

\subsubsection{Achieving Low Latency and Super-Linear Throughput Increase Simultaneously}

For dense models, throughput can be increased by using multiple GPUs and data parallelism (independent replicas with no inter-GPU communication), whereas lower latency can be achieved by techniques like tensor-slicing to partition the model across multiple GPUs \cite{ds-inference-blog}.  The best case scaling in terms of total throughput is linear with respect to the increasing number of GPUs, i.e., a constant throughput per GPU. This is possible for pure data parallel inference scenarios as there is no communication between GPUs.  To reduce latency, tensor-slicing style of model parallelism has proven to be beneficial~\cite{ds-inference-blog} but it comes with the cost --- communication overhead between GPUs --- which often lowers per GPU throughput and results in sublinear scaling of total throughput. In other words, for dense models, we cannot leverage parallelism to optimize both latency and throughput at the same time; there is a tradeoff between them.
MoE inference, however, provides unique opportunities to offer optimized latency and throughput simultaneously while scaling to a large number of devices.

\begin{figure}[htbp]
\centering
\includegraphics[width=0.7\textwidth]{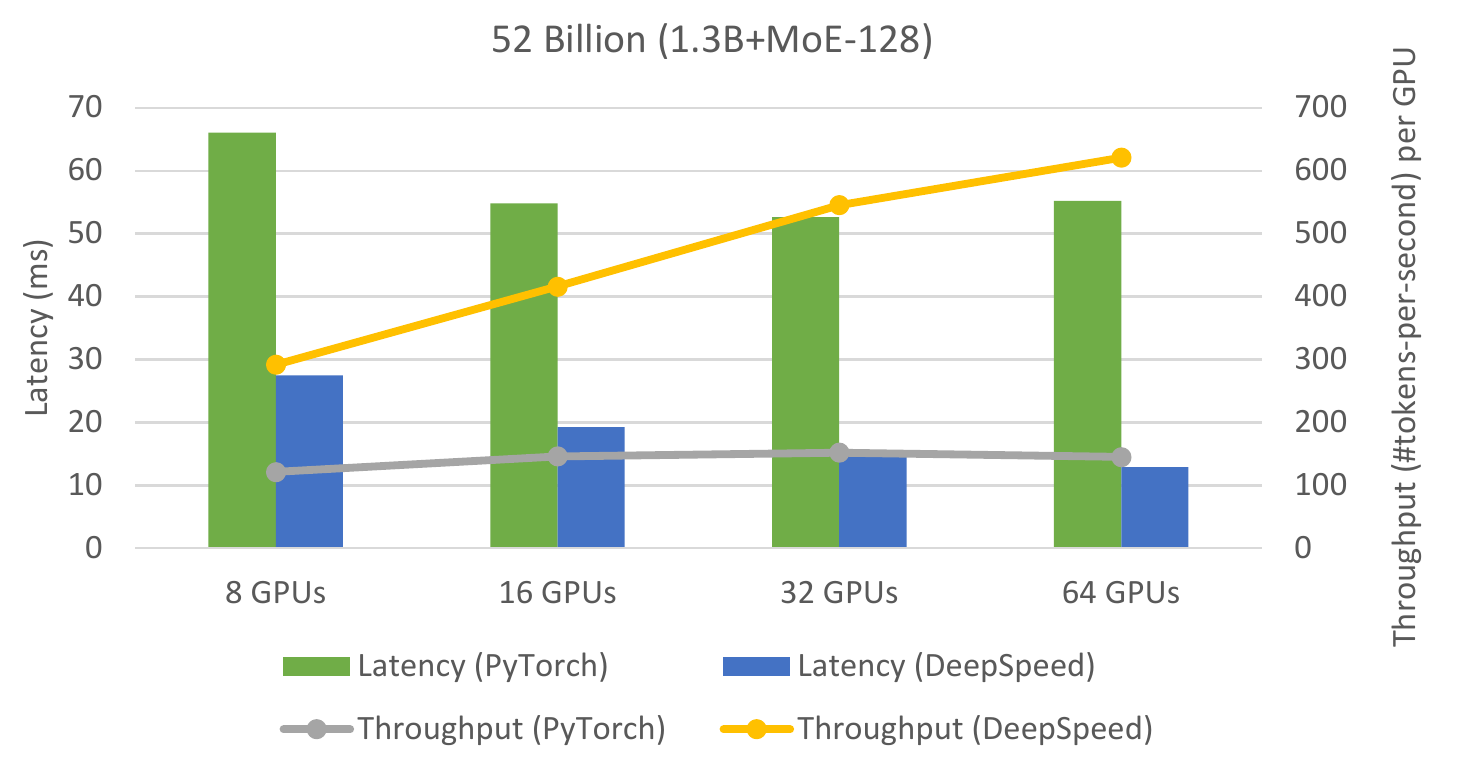}
\caption{Latency and throughput improvement offered by DeepSpeed-MoE inference system (Optimized) over PyTorch (Baseline) for a 52-Billion MoE model with 128 experts, using between 8 to 64 GPUs.}
\label{fig:latency_tput}
\end{figure}

To study these opportunities, we scale a 52B MoE model (1.3B base model and 128 experts) from 8 GPUs to 64 GPUs and observe the latency and throughput trends on DeepSpeed-MoE inference system comparing with a strong baseline: a full-featured distributed PyTorch implementation that is capable of both tensor-slicing and expert-parallelism~\cite{deepspeed-moe-paper}. As shown in Figure~\ref{fig:latency_tput}, we observe that both DeepSpeed and PyTorch reduce the inference latency as we increase the number of GPUs, as expected; although PyTorch is much slower compared to DeepSpeed and only scales up to 32 GPUs. 

The throughput trends, on the other hand, are interesting and merit further analysis. 

As mentioned earlier, the best case throughput scaling for a dense model is linear with respect to the number of GPUs. However, our results in Figure~\ref{fig:latency_tput} show that DeepSpeed obtains increased throughput per GPU when we increase the number of GPUs from 8 to 64 and hence a super-linear increase in total throughput. This is in stark contrast to dense models and shows the major benefit of scaling MoE models over dense models. 

Diving a bit deeper, we see two key properties of expert parallelism at play here: 1) when using expert parallelism, the number of experts per GPU decrease as we increase the number of GPUs. E.g. this 52B MoE model has $128$ total experts; if we serve this using $8$ GPUs, we need $16$ experts per GPU, whereas on $64$ GPUs, we only need $2$ experts per GPU. The decrease in experts per GPU is good for data locality as each GPU is now reading less data (needed for expert parameters) from the memory, and 2) the increase in GPUs could cause performance degradation because of the increased communication between experts (all-to-all) that reside on multiple GPUs. These properties of expert parallelism hold true for both PyTorch and DeepSpeed. However, DeepSpeed-MoE is able to exploit the benefit of expert parallelism to their full potential whereas PyTorch is unable to do so. As DeepSpeed exploits advanced communication optimizations (Section~\ref{sec:opt-alltoall}), it significantly overcomes the communication bottleneck in expert parallelism. At the same time, it has highly optimized kernels (Section~\ref{sec:opt-kernels}) that enable it to take advantage of the increased data locality when experts per GPU are reduced. Both these major wins over PyTorch make DeepSpeed-MoE the framework of choice for getting super-linear increase in throughput with massive scaling, achieving low latency and high throughput simultaneously.

\subsubsection{Low Latency and High Throughput at Unprecedented Scale}

\begin{figure}[htbp]
\centering
\includegraphics[width=\columnwidth]{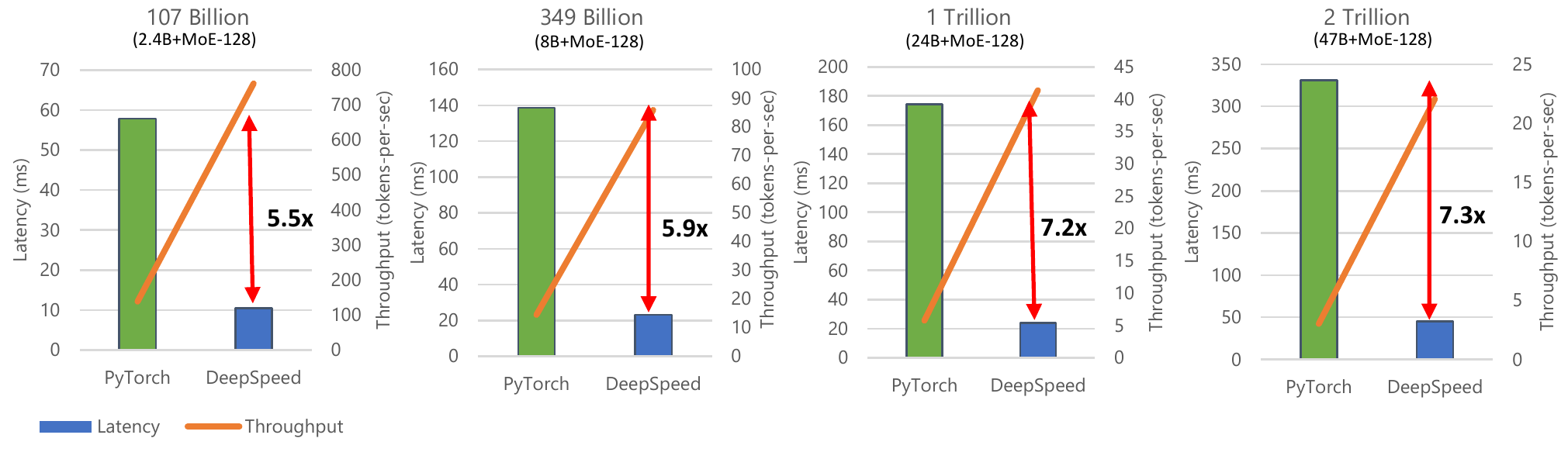}
\caption{Latency and throughput improvement offered by DeepSpeed-MoE (Optimized) over PyTorch (Baseline) for different MoE model sizes (107 billion to 2 trillion parameters). We use 128 GPUs for all configurations for baseline, and 128/256 GPUs for DeepSpeed-MoE (256 GPUs for the trillion-scale models). The throughputs shown here are per GPU and should be multiplied by number of GPUs to get the aggregate throughput of the cluster.}
\label{fig:moe-vs-moe}
\end{figure}

To study the impact of model scale on MoE inference, we now explore MoE models from 107 billion parameters to 2 trillion parameters using PyTorch and DeepSpeed in \fref{fig:moe-vs-moe}.

\begin{itemize}
    \item DeepSpeed-MoE achieves up to 7.3x reduction in latency while achieving up to 7.3x higher throughput compared to the baseline.
    \item By effectively exploiting hundreds of GPUs in parallel, DeepSpeed-MoE achieves an unprecedented scale for inference at incredibly low latencies - a staggering trillion parameter MoE model can be inferenced under 25ms.
\end{itemize}

\subsubsection{Enhanced Benefits of PR-MoE and MoS}

\begin{figure}[htbp]
\centering
\includegraphics[width=0.8\textwidth]{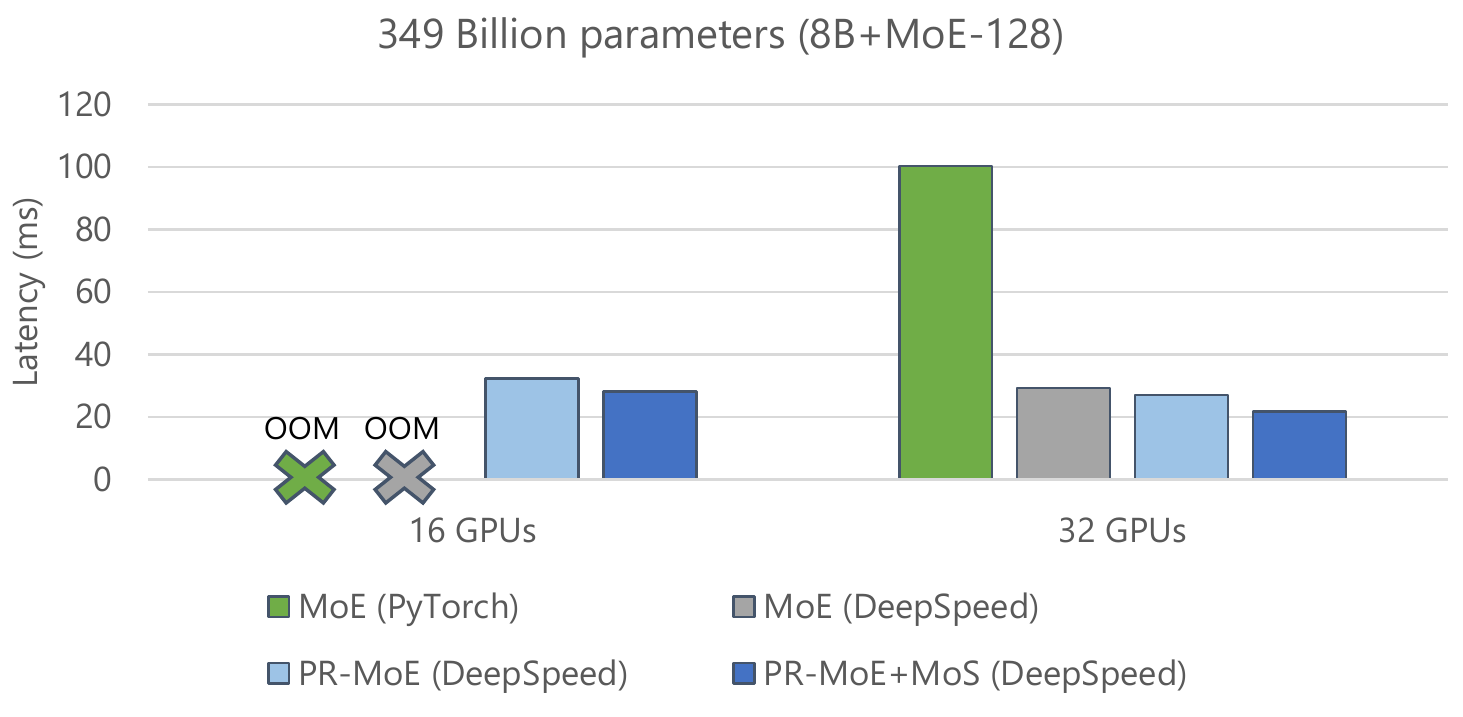}
\caption{2x fewer resources needed for MoE inference when using PR-MoE+MoS. Figure title denotes the model size under standard MoE architecture, and the PR-MoE and PR-MoE+MoS have smaller sizes but same projected quality.}
\label{fig:pr-moe-r1}
\end{figure}

By combining the system optimizations offered by the DeepSpeed-MoE inference system and model innovations of PR-MoE and MoS, DeepSpeed-MoE delivers two more benefits:

(1) Reduce the minimum number of GPUs required to perform inference on these models as shown in Figure \ref{fig:pr-moe-r1}.

\begin{figure}[htbp]
\centering
\includegraphics[width=0.8\textwidth]{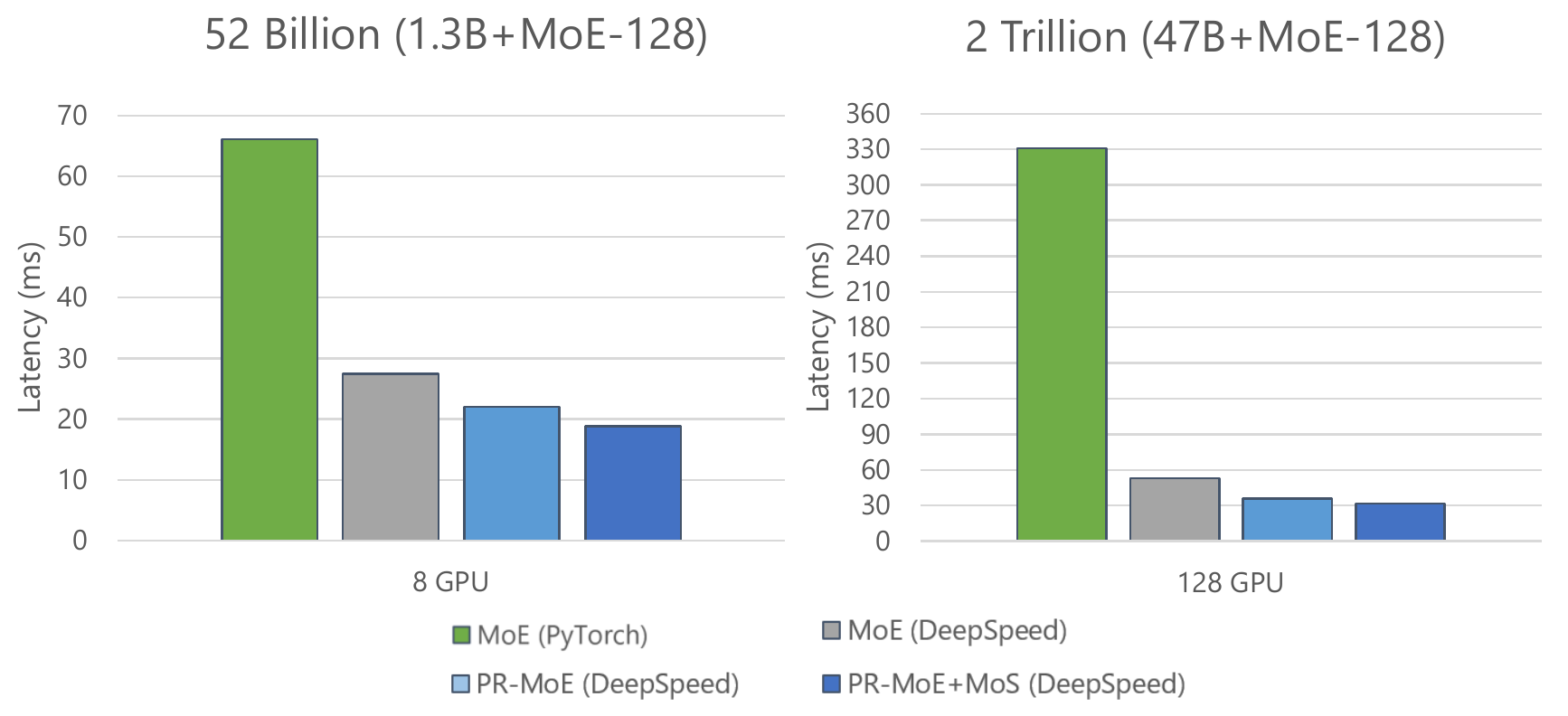}
\caption{Inference latency comparing standard MoE with PR-MoE and PR-MoE+MoS compression on different GPU count and model sizes. Figure title denotes the model size under standard MoE architecture, and the PR-MoE and PR-MoE+MoS have smaller sizes but same quality (projected for the 2 Trillion case).}
\label{fig:pr-moe-r2}
\end{figure}

(2) Further improve both latency and throughput of various MoE model sizes (as shown in~\fref{fig:pr-moe-r2}).

For both Figures \ref{fig:pr-moe-r1} and \ref{fig:pr-moe-r2}, we show a comparison of three model variants along with the baseline version (standard MoE on PyTorch): (i) the standard MoE Model denoted by MoE (DeepSpeed), (ii) the PR-MoE (DeepSpeed), and (iii) the PR-MoE+MoS (DeepSpeed). Results show that the PR-MoE+MoS model offers the lowest latency and enables us to serve the model using only 16 GPUs instead of 32 GPUs. 

\subsubsection{Better Latency and Throughput Than Quality-Equivalent Dense Models}
To better understand the inference performance of MoE models compared to quality-equivalent dense models, it is important to note that although MoE models are 5x faster and cheaper to train, that may not be true for inference. 
Inference performance has different bottlenecks and its primary factor is the amount of data read from memory instead of computation.  

\begin{figure}[htbp]
\centering
\includegraphics[width=0.8\textwidth]{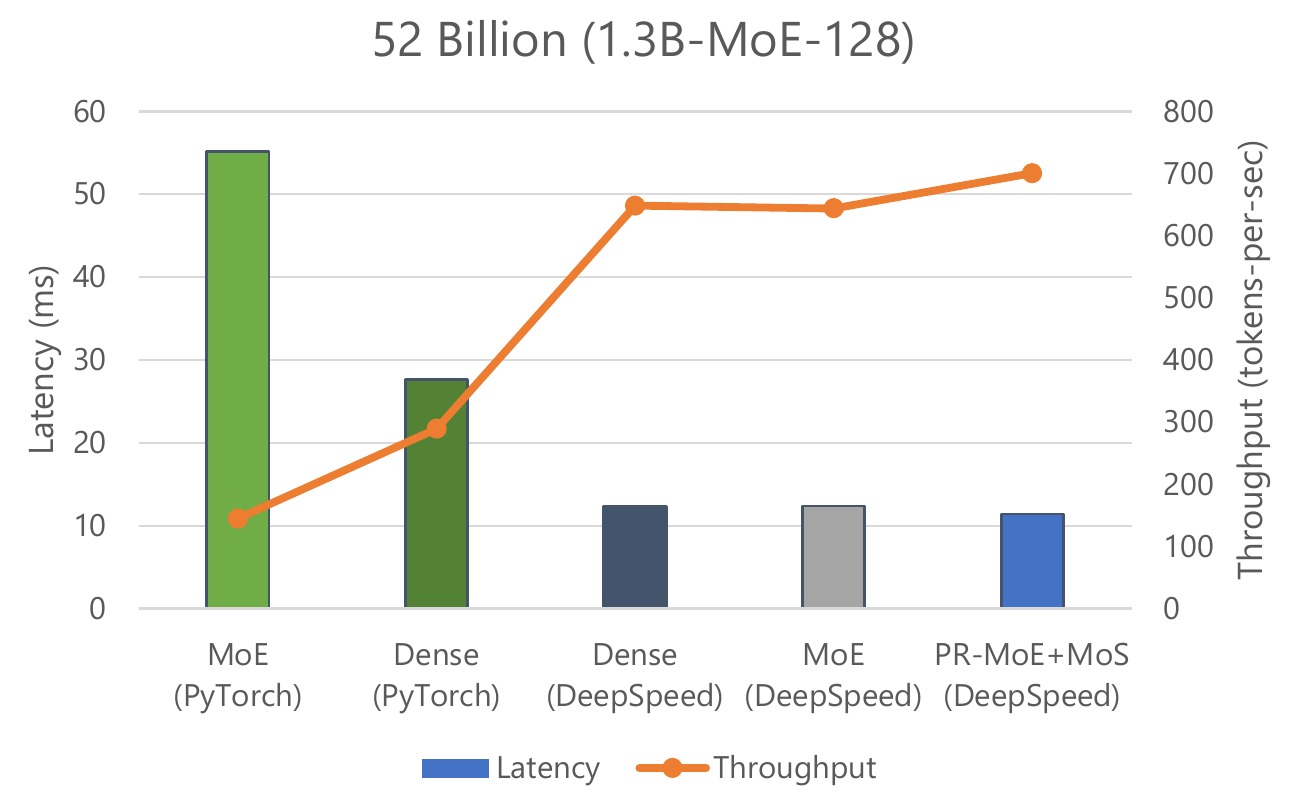}
\caption{Inference latency comparison of MoE models and the quality-equivalent 6.7 billion-parameter dense model. We use 1 GPU for 6.7 billion-parameter model as it offers the lowest latency. We use 128 GPUs for the MoE models. The quality-equivalence has been verified by experiments presented in the training section. Figure title denotes the model size under standard MoE architecture, and the PR-MoE and PR-MoE+MoS have smaller sizes but same quality.}
\label{fig:moe-vs-dense-r1}
\end{figure}

\begin{figure}[htbp]
\centering
\includegraphics[width=0.8\textwidth]{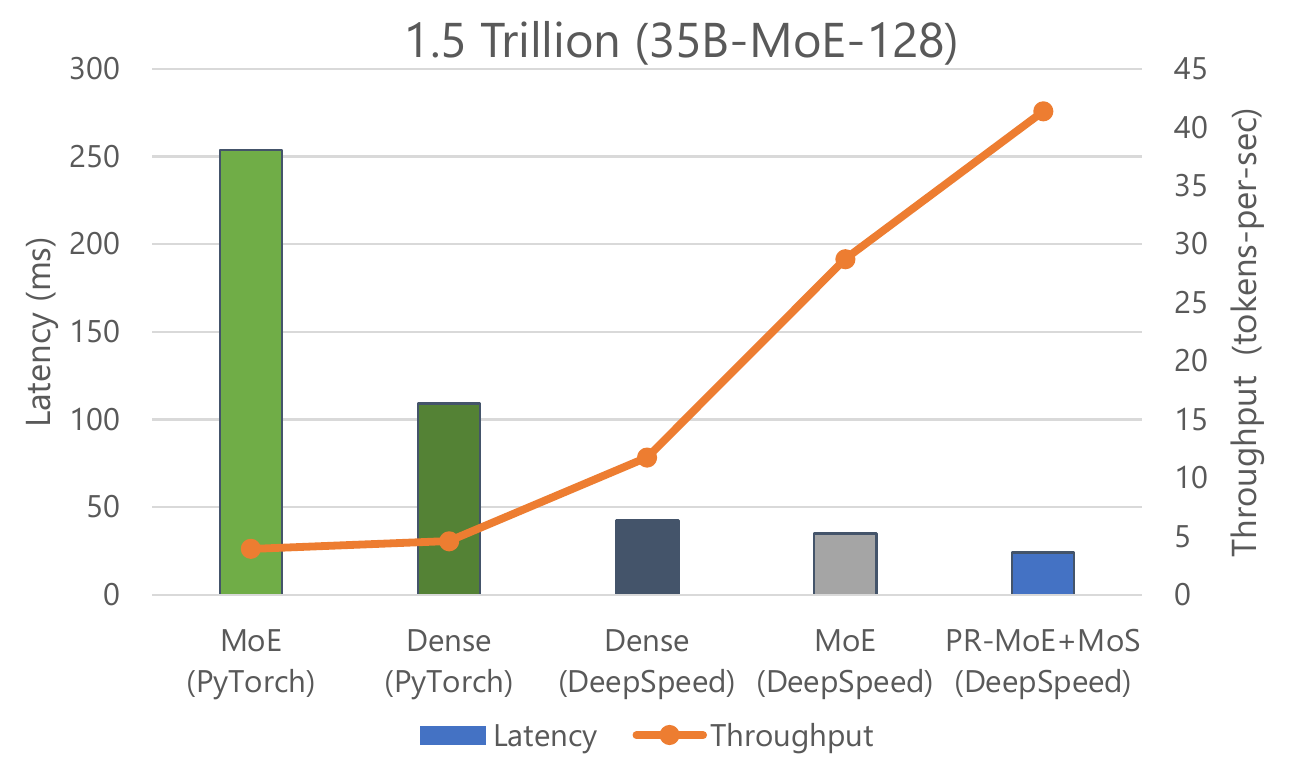}
\caption{Measured inference latency comparison of MoE models and the quality-equivalent 175 billion dense model. We assume the quality equivalence of these two models with the hypothesis that the scaling law of the smaller scale experiments of Figure~\ref{fig:moe-vs-dense-r1} holds, as well as from the observations of the published literature. Figure title denotes the model size under standard MoE architecture, and the PR-MoE and PR-MoE+MoS have smaller sizes but same projected quality.}
\label{fig:moe-vs-dense-r2}
\end{figure}

We show inference latency and throughput for two standard MoE models compared to their quality-equivalent dense models: (1) a 52 billion-parameter MoE model (1.3B-MoE-128) compared to a 6.7 billion-parameter dense model and (2) a 1.5 trillion-parameter MoE model compared to a 175 billion-parameter dense model in~\fref{fig:moe-vs-dense-r1} and \ref{fig:moe-vs-dense-r2}, respectively. We also tested the quality-equivalent PR-MoE+MoS model.

When using PyTorch, MoE model inference is more expensive and slower compared to its quality-equivalent dense models. This is true for both model sizes. 
However, the optimizations in DeepSpeed reverse this trend and make MoE model inference both faster and cheaper compared to quality-equivalent dense models. 
This is a critical result: showing the benefit of MoE models over dense models not only on training but also on inference latency and cost, where real-world deployments care the most.

When comparing the results of~\fref{fig:moe-vs-dense-r1} with~\fref{fig:moe-vs-dense-r2}, we observe that the benefits of MoE models over dense models become even larger with the increase of model size. 
While in ~\fref{fig:moe-vs-dense-r1} the billion-scale PR-MoE+MoS model (served on DeepSpeed-MoE) is 2.4x faster and cheaper than the 6.7 billion-parameter dense model (served on PyTorch), in~\fref{fig:moe-vs-dense-r2} the trillion-scale PR-MoE+MoS model is 4.5x faster and 9x cheaper than the 175 billion-parameter dense model.
The benefits increase for larger models because DeepSpeed leverages parallelism-coordinated optimization to reduce communication overhead when using tensor-slicing on the non-expert part of the model. 
Furthermore, we can take advantage of expert-slicing at this scale, which enables us to scale to a higher number of GPUs compared to the PyTorch baseline. 
In addition, for the larger 1.5 trillion-parameter MoE model, we observed 2x additional improvement in throughput over latency as shown in~\fref{fig:moe-vs-dense-r2}. 
This is because MoE models can run with half the tensor-slicing degree of the dense model (8-way vs. 16-way) and thus two times higher batch size. 

Overall, DeepSpeed-MoE delivers up to 4.5x faster and up to 9x cheaper MoE model inference compared to serving quality-equivalent dense models using PyTorch.
With benefits that scale with model size and hardware resources, as shown from these results, it makes us believe that MoE models will be crucial to bring the next generation of advances in AI scale.

%% file: _s6_conclusion.tex
\section{Looking Forward to the Next Generation of AI Scale}

With the exponential growth of model size recently, we have arrived at the boundary of what modern supercomputing clusters can do to train and serve large models. It becomes harder and harder to achieve better model quality by simply increasing the model size due to insurmountable requirements on hardware resources. The choices we have are to wait for the next generation of hardware or to innovate and improve the training and inference efficiency using current hardware.  

We, along with recent literature \cite{google_glam,artetxe2021efficient}, have demonstrated how MoE-based models can reduce the training cost of the large NLG models by several times compared to their quality-equivalent dense counterparts, offering the possibility to train the next scale of AI models on current generation of hardware. However, prior to this work, to our knowledge, there have been no existing works on how to serve the MoE models (with many more parameters) with latency and cost comparable to or better than the dense models. This is a challenging issue that blocks real-world deployment of large scale MoE models. 

To enable practical and efficient inference for MoE models, we offer novel PR-MoE model architecture and MoS distillation technique to significantly reduce the memory requirements of these models. We also offer an MoE inference framework to achieve incredibly low latency and cost at an unprecedented model scale. Combining these innovations, we are able to make these MoE models not just feasible to serve but able to be used for inference at lower latency and cost than their quality-equivalent dense counterparts. 

As a whole, the new innovations and infrastructures offer a promising path towards training and inference of the next generation of AI scale, without requiring an increase in compute resources. A shift from dense to sparse MoE models can open a path to new directions in the large model landscape, where deploying higher-quality models with fewer resources becomes more widely possible.

%% file: _s7_ack.tex
\section*{Contributions}
\label{sec:contribution}

\noindent\textbf{SR} designed the NLG training experiments and architected the inference system.

\noindent\textbf{CL} led the NLG training experiments (Section~\ref{sec:standard_moe}) and contributed to Section~\ref{sec:moe_parameter_efficiency}.

\noindent\textbf{ZY} led the design and experiments of PR-MoE, and its system support (\sref{sec:pr_moe}).

\noindent\textbf{MZ} led the design and experiments of MoS (Section~\ref{sec:mos}) and memory-efficient checkpointing. 

\noindent\textbf{RYA} and \textbf{AAA} led the development and experiments of the inference system (Section~\ref{sec:optimizing_moe_inference_latency}).

\noindent\textbf{JR} developed, debugged and integrated multiple software features into DeepSpeed.

\noindent\textbf{YH} designed, managed and led the overall research project.

\section*{Acknowledgment}
We thank Olatunji Ruwase from the Microsoft DeepSpeed Team for his contributions on developing, debugging, testing, and releasing the DeepSpeed-MoE software.
This work was done in collaboration with Brandon Norick, Zhun Liu, and Xia Song from the Microsoft Turing Team, Young Jin Kim, Alex Muzio, and Hany Hassan Awadalla from the Microsoft Z-Code Team, and both Saeed Maleki and Madan Musuvathi from the Microsoft SCCL team.